\documentclass[unnumsec,webpdf,contemporary,large]{oup-authoring-template}%

\graphicspath{{Fig/}}

\usepackage{amsmath,amssymb,amsfonts}
\usepackage{longtable}
\usepackage{hyperref}
\usepackage{xurl}

\usepackage{textgreek}

\usepackage{tabularray}
\theoremstyle{thmstyleone}%
\usepackage{mdframed}
\theoremstyle{thmstyletwo}%
\theoremstyle{thmstylethree}%

\begin{document}

\journaltitle{Journal Title Here}
\DOI{DOI HERE}
\copyrightyear{2023}
\pubyear{2019}
\access{Advance Access Publication Date: Day Month Year}
\appnotes{Paper}

\firstpage{1}

%\subtitle{Subject Section}

\title[Fine-Tuning LLMs for Bioinformatics: PRSGPT and BioStarsGPT]{An Empirical Analysis of Fine-Tuning Large Language Models on Bioinformatics Literature: PRSGPT and BioStarsGPT}

\author[1,2,$\ast$]{Muhammad Muneeb}
\author[1,2,$\ast$]{David B. Ascher}

\authormark{Muneeb et al.}

\address[1]{\orgdiv{School of Chemistry and Molecular Biology}, \orgname{The University of Queensland}, \orgaddress{\street{Queen Street}, \postcode{4067}, \state{Queensland}, \country{Australia}}}
\address[2]{\orgdiv{Computational Biology and Clinical Informatics}, \orgname{Baker Heart and Diabetes Institute}, \orgaddress{\street{Commercial Road}, \postcode{3004}, \state{Victoria}, \country{Australia}}}
 
\corresp[$\ast$]{Corresponding authors: David B. Ascher, Email: \href{email:d.ascher@uq.edu.au}{d.ascher@uq.edu.au}}

% \corresp[$\ast$]{Corresponding authors: David B. Ascher, Email: \href{email:d.ascher@uq.edu.au}{d.ascher@uq.edu.au}; Muhammad Muneeb, Email: \href{email:m.muneeb@uq.edu.au}{m.muneeb@uq.edu.au}}

\received{Date}{0}{Year}
\revised{Date}{0}{Year}
\accepted{Date}{0}{Year}

%\editor{Associate Editor: Name}

%\abstract{
%\textbf{Motivation:} .\\
%\textbf{Results:} .\\
%\textbf{Availability:} .\\
%\textbf{Contact:} \href{name@email.com}{name@email.com}\\
%\textbf{Supplementary information:} Supplementary data are available at \textit{Journal Name}
%online.}

\abstract{ 
Large language models (LLMs) often lack specialized knowledge for complex bioinformatics applications. We present a reproducible pipeline for fine-tuning LLMs on specialized bioinformatics data, demonstrated through two use cases: PRSGPT, focused on polygenic risk score (PRS) tools, and BioStarsGPT, trained on community forum discussions. The nine-step pipeline integrates diverse data sources, structured preprocessing, prompt-based question-answer (QA) generation (via Google Gemini), natural language inference (NLI) for quality control, semantic deduplication, clustering-based data splitting, and parameter-efficient fine-tuning using LoRA. We fine-tuned three LLMs (LLaMA-3.2-3B, Qwen2.5-7B, Gemma) and benchmarked them on over 14 lexical and semantic metrics. Qwen2.5-7B emerged as the best performer, with BLEU-4 and ROUGE-1 improvements of 82\% and 70\% for PRSGPT and 6\% and 18\% for BioStarsGPT, respectively. The open-source datasets produced include over 28,000 QA pairs for PRSGPT and 154,282 for BioStarsGPT. Human evaluation of PRSGPT yielded 61.9\% accuracy on the PRS tools comparison task, comparable to Google Gemini (61.4\%), but with richer methodological detail and accurate citations. BioStarsGPT demonstrated 59\% conceptual accuracy across 142 curated bioinformatics questions. Our pipeline enables scalable, domain-specific fine-tuning of LLMs. It enables privacy-preserving, locally deployable bioinformatics assistants, explores their practical applications, and addresses the challenges, limitations, and mitigation strategies associated with their development and use.}
 

\maketitle

\section{Introduction}
Bioinformatics has evolved rapidly due to advances in high-throughput sequencing and the generation of large-scale datasets, making traditional knowledge-sharing methods, such as static documentation, increasingly resource-intensive and inefficient \cite{MorrisonSmith2022,Greene2014,Saparov2025}. In response, LLMs have emerged as powerful tools in biomedical and bioinformatics domains, enabling the extraction of meaningful insights from unstructured data \cite{10433480,llms,pile}. LLMs have been used in clinical decision support \cite{Sarumi2024,nazi2024large,liu2024large}, literature mining and summarization \cite{green2025litsumm,bednarczyk2025scientific}, drug discovery \cite{chakraborty2023artificial}, personalized patient communication \cite{sreenivasanlarge}, large-scale data annotation \cite{Li2024,Anisuzzaman2025}, protein sequence analysis and prediction \cite{Shahid_2023,Schmirler_2024,He2024,Sledzieski_2024,Madani_2023,Chavan}, clinical text processing \cite{Dorfner_2025}, rare disease concept normalization \cite{Wang_2024}, and general biomedical NLP applications \cite{Tinn_2023,Chen2025}.

Recent advances have leveraged the language understanding capabilities of LLMs \cite{lee2020biobert,gu2021domain} to enable researchers to fine-tune models for highly specialized bioinformatics tasks \cite{finetuning}. Fine-tuning protein language models has been shown to significantly boost prediction performance across diverse protein-related tasks \cite{Schmirler_2024,Sledzieski_2024}, while fine-tuning biomedical LLMs improves clinical task performance compared to general-purpose models \cite{Dorfner_2025,Tinn_2023}. A dominant focus within this theme is Parameter-Efficient Fine-Tuning (PEFT), which has emerged as a pivotal approach for adapting large LLMs to new domains despite computational and memory constraints \cite{Huang_2025,Zhang_2024,Xu_2023}. Techniques like LoRA \cite{Tian_2024,hu2022lora} and QLoRA \cite{dettmers2023qlora} enable fine-tuning state-of-the-art LLMs with minimal additional parameters, significantly reducing hardware and storage demands. Together with advances such as retrieval-augmented generation (RAG) and prompt engineering, these methods have broadened the scope of biomedical and bioinformatics applications for fine-tuned LLMs.

Nonetheless, significant gaps remain. Techniques for automated fine-tuning, including the automated selection of hyperparameters and data curation, are still an active area of research \cite{Fang_2024}. Fine-tuning can be highly sensitive to hyperparameters and data quality, making the process prone to instability and inconsistent performance \cite{Dorfner_2025,Shen_2024,Wu_2024,Tinn_2023}. The lack of standardized, comprehensive validation frameworks and benchmarking platforms further complicates rigorous evaluation across bioinformatics tasks \cite{Shen_2024,Chen2025}. Moreover, there is limited literature sharing hands-on experience with fine-tuning LLMs from scratch on domain-specific datasets.

\begin{figure*}[!ht]
    \centering
    \includegraphics[width=0.9\textwidth]{ 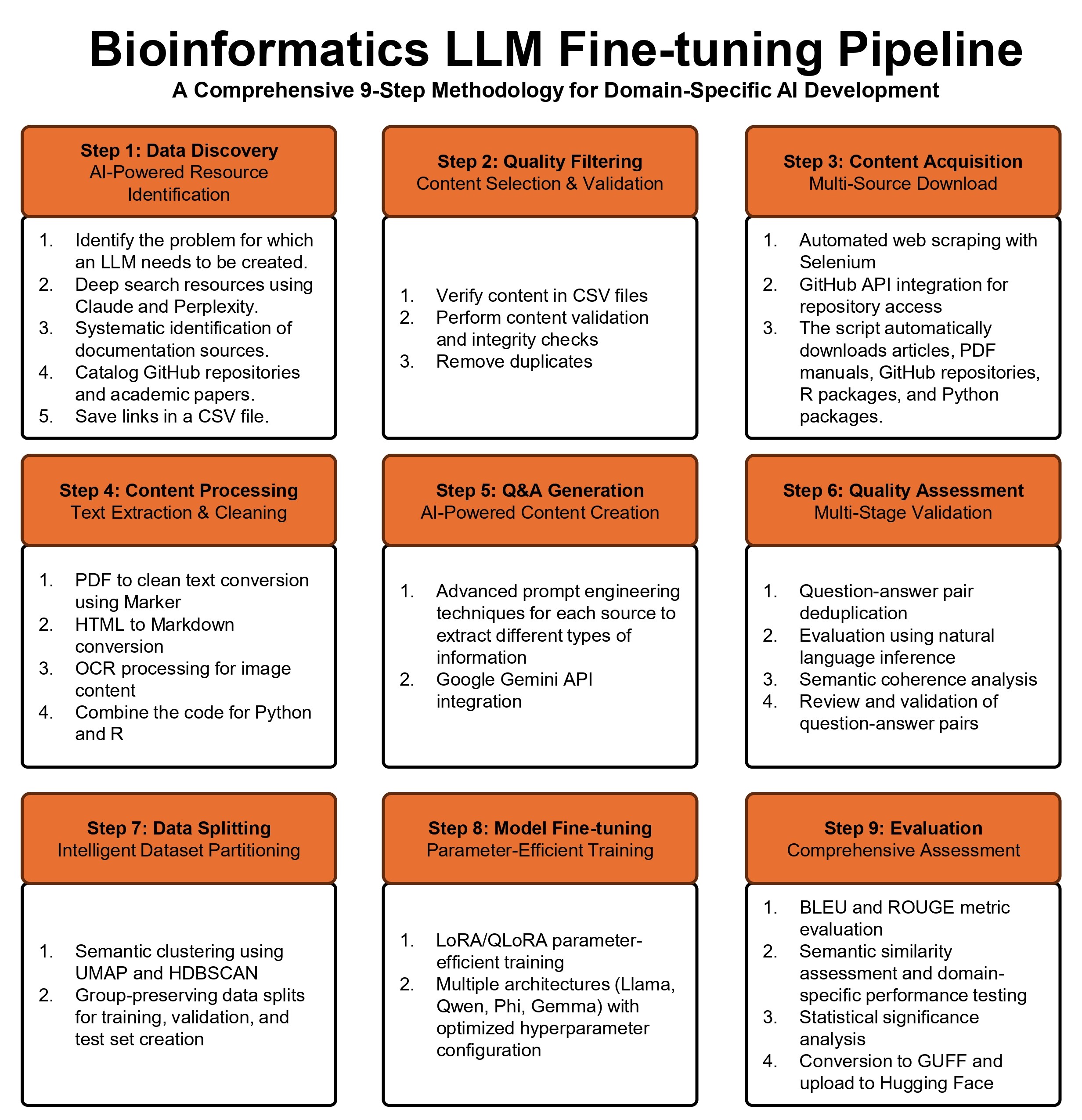}
    \caption{\textbf{Overview of the proposed supervised fine-tuning workflow for bioinformatics applications.} The pipeline consists of data discovery, curation, preprocessing, QA generation, quality assessment, model training, and evaluation, supporting both tool documentation and forum-based data.}
    \label{fig:overview_workflow}
\end{figure*}

% ==================================================
% What this work contributes
% ==================================================
Despite significant advances in applying LLMs within biomedical domains, there is a notable absence of systematic, end-to-end pipelines for fine-tuning LLMs on the diverse, unstructured data sources prevalent in bioinformatics (e.g., tool documentation, research articles, and online forums) \cite{Shahid_2023,Shen_2024,Tinn_2023,Dorfner_2025}. To address this gap, we present a comprehensive, supervised fine-tuning workflow tailored for bioinformatics applications. Our workflow covers all stages of model adaptation, including data discovery, curation, preprocessing, QA generation \cite{Wu_2024}, quality assessment, model training, and evaluation \cite{Shen_2024,Chen2025}. We demonstrate its versatility in two scenarios: \textit{bioinformatics forums} (e.g., the BioStars forum) and \textit{bioinformatics tools} \cite{Hu2022} (e.g., PRS tools) (Figure \ref{fig:overview_workflow}).

The first pipeline leverages deep research features from platforms such as Claude, ChatGPT, and Gemini to (i) identify relevant bioinformatics resources (e.g., articles, GitHub repositories, manuals), (ii) download and convert documents to text, (iii) generate targeted QA pairs via prompt engineering \cite{wei2022chain}, (iv) verify answer quality using a LongFormer model, (v) cluster and partition content for training, validation, and testing, (vi) fine-tune candidate models (Qwen, LLaMA3, Phi, Gemma), (vii) evaluate performance using metrics such as BLEU and METEOR, and (viii) produce a model capable of addressing bioinformatics tool-related queries.

The second pipeline applies a multi-stage process to extract and utilize posts from the BioStars forum (\url{https://biostars.org}). This involves (i) downloading posts, (ii) selecting threads with replies and responses, (iii) generating and verifying QA pairs using AI models, (iv) filtering low-quality answers, (v) splitting data based on semantic clustering, (vi) fine-tuning multiple candidate models, and (vii) conducting performance evaluations.

Together, these pipelines enable researchers to: (i) fine-tune LLMs on custom biomedical datasets, (ii) transform unstructured text into structured QA pairs, and (iii) build literature-based research assistants. They also offer educational and research value. During pipeline development, we examined several factors influencing fine-tuning performance, including model suitability for datasets containing both code and dialogue, optimal parameter configurations based on data complexity and length, the effects of data diversity, and how biases in training data can degrade the performance of fine-tuned models.

\section{Methodology}
\subsection*{PRSGPT: Fine-Tuning Large Language Models on Bioinformatics Resources}
Bioinformatics pipelines rely on complex computational tools, making it challenging for researchers to locate and interpret the information required to run analyses, generate output, and perform calculations using a specific tool \cite{A_Moftah_2018}. General-purpose language models often fail to provide accurate, up-to-date, and tool-specific instructions \cite{_ajewska_2024}. We applied a nine-step fine-tuning pipeline that transforms tool documentation into a structured, high-quality knowledge base. Applying this approach to 32 PRS tools, we extracted, curated, and converted tool-specific information into over 28,000 QA pairs \cite{Lamurias_2020,Ben_Abacha_2019} and fine-tuned the model to provide accurate, tool-specific guidance, serving as a valuable resource for bioinformatics researchers.

The pipeline begins by retrieving links to various information sources related to bioinformatics tools. We used advanced AI-powered search tools, including Claude's Deep Research feature and Perplexity AI, to systematically identify eight content types: Tool PDF documents (official manuals and guides), GitHub repositories (source code and documentation), R and Python package links, published articles (peer-reviewed papers), DOI links, and official website links. All identified resources were cataloged in a structured spreadsheet, available on GitHub: PRSGPT (PRSGPT-Tools-DatasetLinks.xlsx). The downloading process involves the automated extraction of GitHub resources (e.g., README, LICENSE, and dependency files), the use of Selenium-based scraping to extract dynamic website content (up to 500 pages), the retrieval of academic papers using DOIs and preservation of metadata, and the cloning of R and Python package repositories.

The second step focused on transforming raw downloads into clean, structured text. PDF documents were parsed to preserve structure; website HTML pages were converted to Markdown, followed by deduplication and filtering, R package documentation files (.md, .Rmd, .rd) were merged into a single file, and Python files (.py), Jupyter notebooks (.ipynb), and Markdown files were extracted and combined into a consolidated text corpus.

\begin{figure*}[!ht]
    \centering
    \includegraphics[width=1\textwidth]{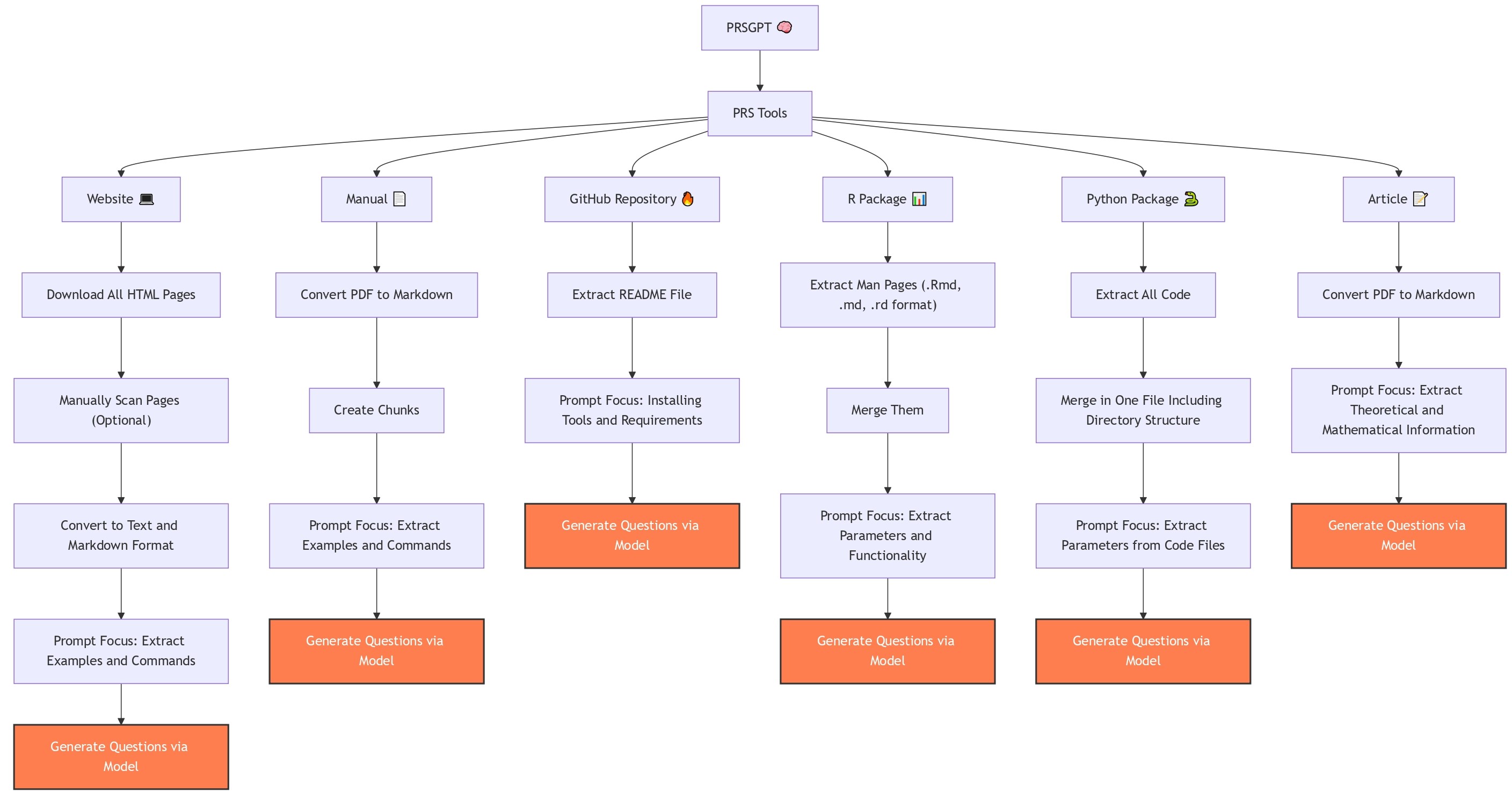}
    \caption{Overview of the pipeline steps for data acquisition and QA generation for Bioinformatics tools.}
    \label{fig:data}
\end{figure*}

The third step used a series of tailored scripts to generate questions and answers for each content type. For GitHub repositories, questions focused on installation, setup, and dependencies. README files provided basic usage and quick start queries. PDF manuals generated detailed questions on tool functionalities and parameter specifications. Published articles focused on theoretical foundations and methodologies. Website content was used to develop practical tutorials. For R packages, we emphasized function usage and environment setup, while Python packages prompted questions on implementation details and hyperparameters. Jupyter notebooks were used to generate questions on interactive workflows and data analysis tutorials. The question generation prompts utilized the Google Gemini model (gemini-2.5-flash-preview-05-20), generating approximately 400 QA pairs per tool (Figure~\ref{fig:data}).

Content verification confirmed file integrity, format consistency, and coverage across eight source types. Question generation succeeded for 100 out of 101 sources (99.0\%), with a 100\% success rate for research articles (26), GitHub READMEs (32), R packages (12), project websites (11), PDF manuals (7), and Jupyter notebooks (2). Most questions originated from project websites (11,503; 40.5\%), followed by Python code (8,579; 30.2\%), GitHub READMEs (2,820; 9.9\%), and research articles (2,151; 7.6\%). R packages (1,259; 4.4\%), PDF manuals (1,225; 4.3\%), and GitHub repositories (841; 3.0\%) contributed smaller shares.

The fifth step merges QA pairs from multiple sources into unified master files for each tool. Exact matching and semantic similarity (utilizing sentence embeddings) \cite{reimers-gurevych-2019-sentence} with a 95\% threshold were employed to eliminate duplicate and near-duplicate questions. This deduplication ensured a non-redundant and informative dataset. The sixth step evaluates the merged datasets using both automated and manual review. Automated filters removed incomplete, irrelevant, or low-quality QA pairs based on answer length and terms such as: "answer does not exist," "source does not contain information," and "except does not contain information."  A transformer-based NLI model \cite{bowman-etal-2015-large,devlin-etal-2019-bert} assessed whether each answer entailed its question, with long answers chunked and scored individually. The Longformer model successfully verified the presence of correct answers for all 1,000 questions within the corresponding output text, achieving 100\% answer coverage with a median score of 60.8\%.

The seventh step uses Sentence-BERT (SBERT) embeddings \cite{reimers-gurevych-2019-sentence} for semantic similarity analysis to cluster questions and evaluate dataset diversity. The data is then stratified and split into training (23,278), validation (5,000), and test (100) sets \cite{neyman1934two}, ensuring balanced representation across clusters while preserving diversity within each subset.

The eighth step measures baseline performance using both general-purpose large language models (e.g., Google Gemini) and local open-source models, including LLaMA (e.g., \texttt{unsloth/Llama-3.2-3B-Instruct}), Phi (\texttt{unsloth/Phi-4}), Qwen (\texttt{unsloth/Qwen2.5-7B}), Mistral (\texttt{unsloth/mistral-7b-instruct-v0.3-bnb-4bit}), and others. The evaluation framework computes a comprehensive suite of metrics: ROUGE (1/2/L) \cite{lin-2004-rouge}, BLEU (1/4) \cite{papineni-etal-2002-bleu}, METEOR \cite{banerjee-lavie-2005-meteor}, Levenshtein similarity \cite{levenshtein1966binary}, Jaccard similarity \cite{jaccard1901etude}, Term Frequency-Inverse Document Frequency (TF-IDF) cosine similarity \cite{sparckjones1972statistical,salton1988term}, SBERT similarity \cite{reimers-gurevych-2019-sentence}, spaCy similarity \cite{honnibal2020spacy}, Word Mover's Distance (WMD) \cite{kusner2015word}, entailment (NLI) \cite{bowman-etal-2015-large}, BERTScore \cite{zhang-etal-2020-bertscore}, CodeBERT similarity \cite{feng-etal-2020-codebert}, and code keyword matching \cite{schleimer2003winnowing}.

The ninth step fine-tunes multiple LLM architectures (LLaMA \cite{touvron2023llama}, Qwen \cite{bai2023qwen}, Gemma \cite{mesnard2024gemma}, Phi \cite{gunasekar2023textbooks}, Mistral \cite{jiang2023mistral}) on the PRS tool dataset using parameter-efficient techniques such as LoRA \cite{hu2022lora} and 4-bit quantization \cite{dettmers2023qlora}. Fine-tuning is performed using the Unsloth library \cite{unsloth} and Hugging Face Trainer \cite{wolf-etal-2020-transformers}, with configurations that include a maximum sequence length of 2048, an adjustable batch size, 50 training epochs, a default learning rate of $2 \times 10^{-4}$, and 4-bit model loading for memory efficiency. Models were trained with evaluation callbacks, early stopping, and detailed logging. This training process yields domain-specific, high-performance models optimized for answering questions in PRS tools. One can aggregate and analyze evaluation results from all models to perform a comparative analysis that identifies the best-performing architectures and highlights their strengths and weaknesses based on key metrics.

\subsection*{BioStarsGPT: Fine-Tuning Large Language Models on Bioinformatics Forums}
Community-driven knowledge platforms, such as BioStars.org, contain expert knowledge and practical insights on bioinformatics topics \cite{Parnell2011}. These platforms host thousands of real-world questions and expert-validated answers, spanning topics from basic sequence analysis to complex genomics pipelines. However, their unstructured nature and high volume of content make it difficult for researchers to efficiently access and utilize this knowledge. We applied the proposed pipeline to fine-tune an LLM on BioStars QA pairs, highlighting the challenges of fine-tuning on such diverse and abundant community resources. These challenges include transforming forum-style data, evaluating training performance, and assessing the feasibility of deployment.

The first step involves collecting question metadata from BioStars.org. For each question, key attributes were extracted, including the Question ID and URL, a cleaned and normalized title, vote count, reply statistics, view counts, other engagement metrics, associated tags, and temporal information such as last updated timestamps. From a total of 90,734 BioStars questions collected over 15 years, RNA-seq (12\%), BLAST (8\%), and genomics (7\%) emerged as the most frequent tags. Around 45\% of questions had no votes, 35\% received one to five votes, and 60\% had at least one reply, indicating moderate engagement. Exploratory data analysis \cite{tukey1977exploratory} (EDA) reveals common title terms, such as "data," "file," and "analysis," indicating a focus on computational workflows. A moderate correlation \cite{pearson1896mathematical} (0.45) between votes and replies suggests that well-received questions prompt more interaction. Most questions were annotated with one to three tags, and a heatmap analysis revealed that votes and views are more closely correlated than replies, highlighting the prevalence of passive content consumption.

The second step involved quality-based filtering, which required each question to have at least one reply, removed malformed or incomplete entries, applied deduplication through semantic similarity detection \cite{salton1975vector}, and enforced a minimum title length to ensure sufficient context. This process improved dataset quality, retaining 45,367 questions out of the initial 90,734. The third step focuses on extracting a complete QA pair by scraping each question's main content and processing embedded multimedia, particularly images, through optical character recognition (OCR) to preserve the bioinformatics discussions. The OCR system achieved an 87\% success rate across 12,543 figures, effectively extracting content such as gene names, sequence annotations, and plot labels. 

The fourth step transforms raw forum discussions into structured QA pairs using Google Gemini (gemini-2.5-pro-exp-03-25) and carefully designed prompt engineering. This process involves generating two complementary question types for each forum post: general overview questions that address the biological context and technical methodology questions that focus on specific tools, parameters, and analysis steps. Prompt templates were constructed to enforce the identification of biological concepts, preserve technical terminology, specify software versions, and maintain accuracy in command-line syntax. Dataset generation and question processing were executed in parallel across 12 threads over approximately one month. The process achieved a 96.3\% valid JSON generation rate, an average semantic similarity of 0.84 ± 0.12 between the generated content and posts, and preserved 92\% of key biological terminology in the first 10,000 questions. EDA of the generated content revealed that the Gemini model successfully generated over 150,000 questions, while Gemini's responses lacked answers for approximately 33,000 questions. Co-occurrence analysis identified strong tag pairings such as \textit{alignment+genome} and \textit{RNA-seq+expression}. A radar plot evaluating content richness showed high scores in lexical diversity and richness of technical terminology. Word frequency analysis followed a power-law distribution, closely resembling Zipf's Law \cite{Piantadosi2014}, indicating the typical "few common, many rare" pattern observed in natural language corpora.

\begin{figure*}[!ht]
    \centering
    \includegraphics[width=\textwidth]{ 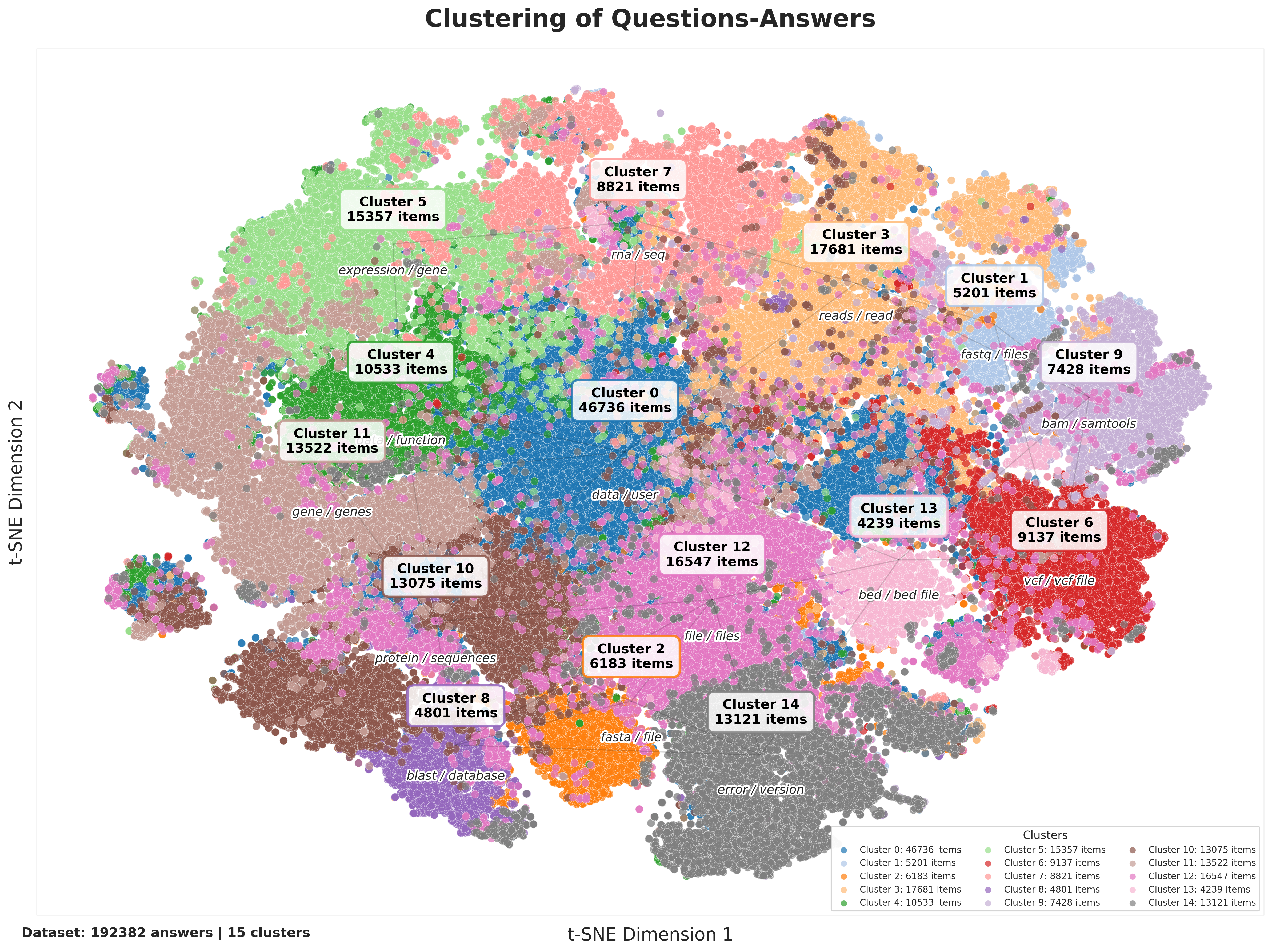}
    \caption{\textbf{This t-SNE visualization clusters 192,382 BioStars answers into 15 distinct topical groups, revealing the semantic organization of bioinformatics discussions on the platform.} The largest cluster (Cluster 0, 46,736 items) focuses on general data/user themes. In contrast, others correspond to specialized workflows such as RNA sequencing, genomic coordinates (BED), variant calling (VCF), and sequence alignment (BAM/SAMtools). The spatial separation of clusters reflects forum diversity.}
    \label{fig:clustering}
\end{figure*}

The fifth step combines individual QA pairs into a unified dataset and assesses topic diversity through a clustering-based analysis. To visualize semantic structure, we applied t-distributed stochastic neighbor embedding (t-SNE) dimensionality \cite{vandermaaten08a} reduction on Sentence-BERT embeddings, revealing 15 major clusters (Figure~\ref{fig:clustering}).  

The sixth step involves filtering based on answer quality, discarding incomplete or irrelevant responses, and enforcing a 2048-token length limit for LLM compatibility. To assess QA alignment, NLI was performed using Longformer and DeBERTa-v3-large on the first 10,000 questions, evaluating all pairs with average entailment scores of 54.90 for general and 55.01 for technical questions. After filtering, the dataset was reduced from 193,133 to 154,282 high-quality QA pairs, resulting in an 80.1\% retention rate.

To prevent data leakage and ensure balanced partitions, a clustering-based data splitting \cite{dhillon2001concept} strategy was applied. This approach preserved content integrity while maintaining diversity across training, validation, and test sets, similar to PRSGPT data split. For this dataset, we ensured that both general and technical questions derived from the same post were assigned to the same set. The final dataset comprises 154,282 QA pairs, with 93.1\% (144,137 samples) allocated to training, 6.5\% (10,001 samples) to validation, and 0.4\% (100 samples) reserved for testing.  

The eighth and ninth steps involved evaluating model performance on the test set using both general-purpose models, such as Google Gemini, and locally hosted models. Subsequently, four selected models were fine-tuned on the curated dataset. Final evaluations employed the same metrics and methodology as the PRSGPT pipeline, allowing direct comparison between baseline and fine-tuned model performance.

\section{Results}
\label{results}
After dataset generation, we fine-tuned four large language models—LLaMA-3.2-3B, Qwen2.5-7B, Phi-4, and Gemma-2-27B—using a parameter-efficient strategy based on the UnSloth fine-tuning scripts \url{https://github.com/unslothai/unsloth} \cite{unsloth} and fine-tuning employed a LoRA rank of 16, which determines the dimensionality of the low-rank matrices inserted into the attention layers, enabling efficient adaptation with minimal trainable parameters. A scaling factor $\alpha = 32$ and a dropout rate of 0.1 were used to balance adaptation and reduce overfitting. Training proceeded for up to 50 epochs, with early stopping based on validation loss, using a cosine learning rate schedule that started at $2 \times 10^{-4}$. A batch size of 16 with gradient accumulation was used to accommodate hardware limitations. Optimization used 8-bit AdamW with gradient clipping for stability. All models were trained with a context window of 2048 tokens. Inference used consistent decoding settings: a maximum of 1024 new tokens, temperature of 0.7, top-$p$ sampling at 0.95, and top-$k$ set to 50.

Model performance was evaluated after each epoch, and results from the final checkpoint were reported. The same training setup was applied to both PRSGPT and BioStarsGPT pipelines.

\subsection{PRSGPT}
The dataset was partitioned into training, validation, and test sets, comprising 23,278 training samples, 5,000 validation samples, and 100 test samples. Baseline performance was evaluated for five large language models—Gemini-2.5, LLaMA-3.2-3B, Phi-4, Qwen2.5-7B, and Mistral-7B—without fine-tuning. The output of each model on the test set was compared against the ground-truth responses. Table~\ref{tab:baselineprsgpt} summarizes the evaluation results.
Mistral-7B performed best overall, with strong semantic coherence and lexical overlap. Qwen2.5-7B led in ROUGE and TF-IDF, LLaMA-3.2-3B in code keyword matching, and Phi-4 in logical entailment. Gemini-2.5 excelled in entailment but underperformed in other metrics, likely due to providing excessively detailed responses that exceeded the expected scope.

\begin{table*}[!ht]
\centering
\begin{tabular}{|l|c|c|c|c|c|}
\hline
\textbf{Metric} & \textbf{Gemini-2.5} & \textbf{LLaMA-3.2-3B} & \textbf{Phi-4} & \textbf{Qwen2.5-7B} & \textbf{Mistral-7B} \\
\hline
Exact Match            & 0.00 & 0.00 & 0.00 & 0.00 & 0.00 \\
Levenshtein Similarity & 0.11 & 0.21 & 0.19 & 0.21 & \textbf{0.22} \\
Jaccard Similarity     & 0.09 & 0.18 & 0.16 & \textbf{0.19} & 0.19 \\
TF-IDF Cosine          & 0.20 & 0.38 & 0.27 & \textbf{0.39} & 0.39 \\
ROUGE-1                & 0.10 & 0.23 & 0.19 & \textbf{0.24} & 0.23 \\
ROUGE-2                & 0.03 & 0.07 & 0.05 & \textbf{0.08} & 0.07 \\
ROUGE-L                & 0.09 & 0.21 & 0.18 & \textbf{0.23} & 0.22 \\
BLEU-1                 & 0.04 & 0.21 & 0.18 & 0.20 & \textbf{0.23} \\
BLEU-4                 & 0.01 & 0.05 & 0.04 & 0.05 & \textbf{0.05} \\
METEOR                 & 0.06 & 0.23 & 0.22 & 0.24 & \textbf{0.26} \\
SpaCy Similarity       & 0.80 & 0.93 & 0.93 & 0.92 & \textbf{0.94} \\
SBERT Similarity       & 0.40 & 0.71 & 0.61 & 0.73 & \textbf{0.73} \\
WMD Similarity         & 0.48 & 0.56 & 0.54 & 0.56 & \textbf{0.56} \\
Entailment Score       & 0.50 & 0.50 & \textbf{0.51} & 0.49 & 0.50 \\
\hline
\end{tabular}
\caption{Zero-shot performance of unfine-tuned large language models on the PRSGPT test set. Comparison of five models across lexical, semantic, and entailment metrics. The highest score per metric is highlighted in \textbf{bold}.}
\label{tab:baselineprsgpt}
\end{table*}

After evaluating the performance of the unfine-tuned models, we fine-tuned three models—Gemma, LLaMA-3.2-3B, and Qwen2.5-7B—for up to 7 days each, with early stopping applied when no further improvement was observed. The models completed 18, 47, and 37 epochs, respectively. After each epoch, inference was conducted on the test set by prompting the models with test questions and comparing their responses to the ground-truth answers. Final metrics from the last epoch were reported and compared to the best results of the corresponding unfine-tuned models.

\begin{table*}[!ht]
\centering
\begin{tabular}{|l|c|c|c|c|c|c|c|}
\hline
\textbf{Metric} & \textbf{BUT} & \textbf{L3B} & \textbf{Q7B} & \textbf{GM} & \textbf{L3B G/L} & \textbf{Q7B G/L} & \textbf{GM G/L} \\
\hline
Exact Match         & 0.00 & 0.00 & \textbf{0.41} & 0.00 & 0.00 & \textbf{0.41} & 0.00 \\
Levenshtein Similarity & 0.22 & 0.30 & \textbf{0.93} & 0.16 & 0.08 & \textbf{0.71} & -0.07 \\
Jaccard Similarity     & 0.19 & 0.28 & \textbf{0.91} & 0.24 & 0.09 & \textbf{0.72} & 0.05 \\
TF-IDF Cosine          & 0.39 & 0.44 & \textbf{0.95} & 0.37 & 0.04 & \textbf{0.55} & -0.02 \\
ROUGE-1                & 0.24 & 0.42 & \textbf{0.94} & 0.24 & 0.17 & \textbf{0.70} & -0.01 \\
ROUGE-2                & 0.08 & 0.17 & \textbf{0.92} & 0.09 & 0.10 & \textbf{0.84} & 0.02 \\
ROUGE-L                & 0.23 & 0.27 & \textbf{0.93} & 0.14 & 0.04 & \textbf{0.70} & -0.09 \\
BLEU-1                 & 0.23 & 0.37 & \textbf{0.91} & 0.22 & 0.14 & \textbf{0.68} & -0.02 \\
BLEU-4                 & 0.05 & 0.13 & \textbf{0.87} & 0.08 & 0.07 & \textbf{0.82} & 0.03 \\
METEOR                 & 0.26 & 0.35 & \textbf{0.93} & 0.32 & 0.09 & \textbf{0.67} & 0.05 \\
spaCy Similarity       & 0.94 & 0.94 & \textbf{1.00} & 0.95 & 0.00 & \textbf{0.06} & 0.01 \\
SBERT Similarity       & 0.73 & 0.79 & \textbf{0.98} & 0.78 & 0.06 & \textbf{0.26} & 0.05 \\
WMD Similarity         & 0.56 & 0.62 & \textbf{0.96} & 0.60 & 0.06 & \textbf{0.40} & 0.04 \\
Entailment             & \textbf{0.51} & 0.23 & 0.34 & 0.00 & -0.28 & -0.17 & \textbf{-0.51} \\
\hline
\end{tabular}
\caption{Comparison of fine-tuned models against the best-performing unfine-tuned model (BUT) across multiple evaluation metrics. The table includes scores for LLaMA-3.2-3B (L3B), Qwen2.5-7B (Q7B), and Gemma (GM), alongside their respective gains or losses (G/L) relative to the best unfine-tuned baseline.}
\label{tab:fine_tuned_comparison}
\end{table*}

Table~\ref{tab:fine_tuned_comparison} compares fine-tuned models LLaMA-3.2-3B (L3B), Qwen2.5-7B (Q7B), and Gemma (GM) against the best unfine-tuned model (BUT) across lexical and semantic metrics on the PRSGPT test set. Qwen2.5-7B showed the most considerable improvements in metrics such as BLEU-4 (+0.82), ROUGE-1 (+0.70), and SBERT Similarity (+0.26). LLaMA-3.2-3B had moderate, often significant gains, while Gemma's improvements were marginal and not statistically significant ($p > 0.05$). These findings indicate that fine-tuning, especially for Qwen2.5-7B, substantially enhances performance. The limited gains for LLaMA and Gemma likely reflect their difficulty in handling the mixed dialogue and code responses present in the training data.

To evaluate the factual accuracy of responses generated by PRSGPT, we used a reference table from Ma et al.~\cite{Ma2021}, which presents a comprehensive comparison of PRS tools based on various methodological parameters. This table, provided as Supplementary Material Table 1 in their paper, includes detailed information on eleven attributes such as input data type, annotation usage, linkage disequilibrium (LD) matrix modeling, validation strategy, effect size modeling assumptions, effect size and variance distributions, inference algorithms, software packages, and programming languages. Using this external resource as ground truth, we generated 176 questions related to sixteen tools, covering these eleven attributes. Each response from PRSGPT was scored using a three-point rubric: Correct (1.0), Partially Correct (0.5), and Incorrect (0.0). PRSGPT achieved 109 correct responses (61.9\%), 11.4\% partially correct, and 26.7\% incorrect, a performance closely comparable to that of Google Gemini.

For benchmarking, we also evaluated Claude and Google Gemini on the same set of 176 questions. Claude achieved the highest accuracy with 126 correct responses (71.6\%), followed by Google Gemini with 108 correct responses (61.4\%). Despite slightly lower quantitative accuracy, PRSGPT demonstrated several qualitative advantages, including the generation of detailed methodological descriptions, such as mathematical equations for effect size estimation, and the inclusion of accurate source citations supporting its answers. Unlike Claude, which tended to provide high-level summaries, PRSGPT offered more granular explanations, increasing its value for scientific research. These outputs, including model responses, are publicly available on GitHub in the file \texttt{prstools\_output.pdf}.

Nonetheless, PRSGPT exhibited certain limitations. It struggled with comparative questions, such as generating tables comparing input formats across tools, likely because it was fine-tuned primarily on tool-specific queries rather than comparative tasks. The model was also case-sensitive, interpreting tool names differently depending on capitalization and occasionally hallucinating responses when the tool name was missing from the prompt. Compared to BioStarGPT, which often failed to provide accurate source links, PRSGPT was more reliable in citing valid references. A significant limitation was observed with 4-bit quantization, which reduced the model size from 15 GB to 5 GB but substantially degraded response accuracy and source citation quality.

Another challenge arose from tools with similar names—for example, LDPred and LDPred2—which PRSGPT sometimes confused unless explicitly distinguished during fine-tuning or via prompt engineering. Additionally, the manuals and documentation used during training sometimes referenced other tools, which complicated evaluation. For example, a question about LDPred might be answered from an LDPred2 manual that mentions other software, leading PRSGPT to incorrectly state that the information was unavailable. These issues emphasize the importance of carefully curating training datasets and ensuring that both training and evaluation questions are precisely scoped and unambiguous.

Despite these limitations, PRSGPT remains a valuable tool for generating detailed, referenced outputs and can be effectively used for various scientific and research-related tasks involving PRS methodologies and comparative analysis (GitHub: PRSGPT (tool\_eval\_output.pdf)).

\subsection*{BioStarsGPT}
The BioStarsGPT dataset was split into training (144,137 samples), validation (10,001 samples), and test (100 samples) sets. We evaluated the baseline performance of five models without fine-tuning by comparing their outputs on the test set against ground-truth answers. Table~\ref{tab:baseline_biostargpt_rounded} summarizes the zero-shot performance across multiple lexical and semantic metrics.

\begin{table*}[!ht]
\centering
\begin{tabular}{|l|c|c|c|c|c|}
\hline
\textbf{Metric} & \textbf{Gemini-2.5} & \textbf{LLaMA-3.2-3B} & \textbf{Phi-4} & \textbf{Qwen2.5-7B} & \textbf{Mistral-7B} \\
\hline
Exact Match            & 0.00 & 0.00 & 0.00 & 0.00 & 0.00 \\
Levenshtein Similarity & 0.06 & 0.22 & 0.21 & 0.23 & \textbf{0.25} \\
Jaccard Similarity     & 0.08 & 0.19 & 0.16 & 0.19 & \textbf{0.19} \\
TF-IDF Cosine          & 0.16 & 0.41 & 0.33 & 0.42 & \textbf{0.43} \\
ROUGE-1                & 0.08 & 0.25 & 0.21 & \textbf{0.26} & \textbf{0.26} \\
ROUGE-2                & 0.00 & 0.06 & 0.05 & \textbf{0.07} & 0.06 \\
ROUGE-L                & 0.06 & 0.23 & 0.20 & \textbf{0.24} & \textbf{0.24} \\
BLEU-1                 & 0.00 & 0.23 & 0.20 & 0.24 & \textbf{0.27} \\
BLEU-4                 & 0.00 & 0.04 & 0.03 & 0.04 & \textbf{0.04} \\
METEOR                 & 0.04 & 0.27 & 0.23 & 0.26 & \textbf{0.27} \\
spaCy Similarity       & 0.87 & 0.93 & 0.91 & 0.94 & \textbf{0.95} \\
SBERT Similarity       & 0.28 & 0.74 & 0.65 & 0.74 & \textbf{0.76} \\
WMD Similarity         & 0.47 & 0.57 & 0.55 & 0.57 & \textbf{0.57} \\
Entailment Score       & 0.51 & \textbf{0.54} & 0.48 & 0.48 & 0.49 \\
\hline
\end{tabular}
\caption{Zero-shot performance of unfine-tuned large language models on the BioStarsGPT test set. The highest score per metric is highlighted in \textbf{bold}.}
\label{tab:baseline_biostargpt_rounded}
\end{table*}

As shown in Table~\ref{tab:baseline_biostargpt_rounded}, all models had 0\% exact match accuracy, underscoring the challenge of exact replication without fine-tuning. Mistral-7B scored highest on most lexical and semantic metrics, while Qwen2.5-7B led in ROUGE scores. Gemini-2.5 generally underperformed.  

For fine-tuning, we ran 2, 15, and 4 epochs for Gemma, LLaMA-3.2-3B, and Qwen2.5-7B, respectively, reporting results from the final epoch. The training dataset included about 144,137 questions, and fine-tuning was limited to seven days. Training the Phi-4 model could not be completed due to extended runtimes and system crashes. LLaMA-3.2-3B showed initial loss reduction, but loss later increased, resulting in no meaningful performance improvement. Limited resources prevented longer training. In contrast, Qwen2.5-7B showed good progress but would need more than seven days to fully fine-tune.
\begin{table*}[!ht]
\centering
\begin{tabular}{|l|c|c|c|c|c|c|c|}
\hline
\textbf{Metric} & \textbf{BUT} & \textbf{L3B} & \textbf{Q7B} & \textbf{GM} & \textbf{L3B G/L} & \textbf{Q7B G/L} & \textbf{GM G/L} \\
\hline
Exact Match       & 0.00 & 0.00 & 0.00 & 0.00 & 0.00 & 0.00 & 0.00 \\
Levenshtein Sim   & 0.25 & 0.27 & \textbf{0.30} & 0.17 & 0.02 & \textbf{0.05} & -0.08 \\
Jaccard Sim       & 0.19 & 0.22 & \textbf{0.26} & 0.19 & 0.03 & \textbf{0.07} & 0.00 \\
TF-IDF Cosine     & 0.43 & 0.39 & \textbf{0.45} & 0.40 & -0.04 & \textbf{0.02} & -0.03 \\
ROUGE-1           & 0.26 & 0.39 & \textbf{0.44} & 0.25 & 0.13 & \textbf{0.18} & -0.01 \\
ROUGE-2           & 0.07 & 0.12 & \textbf{0.16} & 0.08 & 0.05 & \textbf{0.09} & 0.01 \\
ROUGE-L           & 0.24 & 0.22 & \textbf{0.25} & 0.13 & -0.02 & \textbf{0.01} & -0.11 \\
BLEU-1            & 0.27 & 0.37 & \textbf{0.39} & 0.16 & 0.10 & \textbf{0.12} & -0.11 \\
BLEU-4            & 0.04 & 0.08 & \textbf{0.10} & 0.04 & 0.04 & \textbf{0.06} & 0.00 \\
METEOR            & 0.27 & 0.24 & \textbf{0.27} & 0.27 & -0.03 & \textbf{0.00} & 0.00 \\
spaCy Sim         & 0.95 & 0.95 & \textbf{0.97} & 0.97 & 0.00 & \textbf{0.02} & 0.02 \\
SBERT Sim         & 0.76 & 0.74 & \textbf{0.77} & 0.74 & -0.02 & \textbf{0.01} & -0.02 \\
WMD Sim           & 0.57 & 0.59 & \textbf{0.63} & 0.60 & 0.02 & \textbf{0.06} & 0.03 \\
Entailment        & 0.49 & 0.15 & \textbf{0.18} & 0.00 & -0.34 & \textbf{-0.31} & -0.49 \\
\hline
\end{tabular}
\caption{Comparison of fine-tuned models against the best-performing unfine-tuned baseline (BUT) across selected lexical and semantic metrics. Abbreviations: BUT = Best Unfine-Tuned, L3B = LLaMA-3.2-3B, Q7B = Qwen2.5-7B, GM = Gemma, G/L = Gain or Loss relative to BUT. Bold values indicate the greatest performance gains.}
\label{tab:fine_tuned_vs_unft}
\end{table*}

Table~\ref{tab:fine_tuned_vs_unft} compares three fine-tuned models—LLaMA-3.2-3B (L3B), Qwen2.5-7B (Q7B), and Gemma (GM)—with the best-performing unfine-tuned baseline (BUT) on 14 evaluation metrics. Qwen2.5-7B consistently outperformed the baseline, showing significant improvements in lexical (e.g., BLEU-4: +0.06, ROUGE-1: +0.18) and semantic similarity (e.g., SBERT: +0.01, WMD: +0.06). LLaMA-3.2-3B showed moderate gains in most lexical metrics (e.g., ROUGE-L: +0.01, BLEU-1: +0.10), though some were not statistically significant ($p > 0.05$). Gemma exhibited minimal gains, with no significant improvements. These findings suggest that fine-tuning, especially of Qwen2.5-7B, yields substantial and statistically reliable performance enhancements.

To evaluate the performance of \texttt{BioStarsGPT}, we conducted a human assessment using a diverse set of questions randomly selected from the BioStars forum. The model produced six accurate answers, seven partially correct responses, one incorrect answer, and one that could not be verified. Detailed results, including question references and responses, are available on GitHub: BioStarsGPT (BioStarsGPT-HumanTesting.xlsx). The model successfully generated accurate commands, referenced tools such as \texttt{bcftools}, and cited relevant documentation like LDpred panels. In several instances, its responses matched or surpassed user-provided answers, with seven verified as correct by ChatGPT. However, seven answers were only partially accurate, often due to vague commands, missing documentation, or misinterpretations of concepts such as the null hypothesis in the context of PRS. Occasionally, the model closely replicated forum content without improving clarity or interpretive depth.

A key factor influencing performance was the quality and scope of the training data. Many BioStars questions and answers were brief or incomplete, which the model reflected in its minimal responses. When faced with poorly developed discussion threads, it struggled to expand or integrate related information. Although it provided more detailed explanations when asked for, its ability to generalize beyond training examples remained limited. This limitation became evident when compared to outputs from Google Gemini Pro 2.5, which enhanced initial answers into more comprehensive, technically accurate tutorials, several of which are available on GitHub. Notably, guiding Gemini Pro 2.5 with outputs from BioStarsGPT improved responses, highlighting the benefit of prompt engineering and multi-model collaboration in refining domain-specific answers.

Further evaluation involved 142 questions spanning four bioinformatics domains: PRS, gene expression, pathway analysis, and genotype-phenotype prediction. These questions covered tool usage, statistical methods, data preprocessing, and visualization. Google Gemini Pro 2.5's review of the responses generated from BioStarsGPT found that 84 responses (59\%) were conceptually sound but required technical refinement, while 58 (41\%) were incorrect or irrelevant. The model sometimes used phrasing or terminology that did not align precisely with the questions and occasionally combined information from unrelated topics. Its limited ability to provide consistent references and maintain multi-turn dialogue reduces its effectiveness for complex bioinformatics tasks. Despite these limitations, BioStarsGPT performed well on foundational bioinformatics questions related to GWAS, data formats, and basic concepts, likely reflecting its training on frequently asked, well-structured queries. Overall, we rate BioStarsGPT's performance as \textbf{3 out of 5}.

\section{Conclusion}
\label{discussion}
This study presents a comprehensive pipeline for fine-tuning LLMs on bioinformatics data. We tested the proposed pipeline on two use cases: PRSGPT, which focuses on polygenic risk score tool comparisons, and BioStarsGPT, which addresses diverse bioinformatics queries from community forums \cite{Li2024,Simon2024}.

Our nine-step pipeline synthesizes diverse bioinformatics sources—such as PDF documents, GitHub repositories, research articles, websites, and forum discussions into structured QA pairs. We used NLI \cite{Yang2020}, semantic deduplication, and clustering-based data splitting to divide the dataset. We then fine-tuned LLMs on these datasets and tested them across fourteen evaluation metrics. Finally, we conducted a human evaluation to assess the effectiveness of the proposed pipeline.

Several technical challenges emerged during pipeline development, which we systematically addressed. \textbf{I. Data Collection:} Diverse sources such as PDF documents, GitHub repositories, research articles, websites, and forums required extensive web scraping and custom extraction scripts. \textbf{II. Preprocessing:} We implemented multi-stage pipelines with batch processing to efficiently manage large, complex datasets. \textbf{III. Mixed Content Handling:} The presence of both natural language and executable code demanded specialized prompt engineering and expanded evaluation metrics to properly represent data and assess the model's performance. \textbf{IV. Data Splitting:} To prevent leakage, we applied semantic clustering techniques that kept related content grouped during dataset partitioning and ensured that the test set was selected from all diverse clusters. \textbf{V. Quality Control:} Automated NLI validation with a Longformer model was employed to ensure that the generated content aligned with the source text \cite{Keloth2024,Dessimoz2024}. \textbf{VI. Training Stability:} We mitigated instabilities such as overfitting and gradient fluctuations by employing early stopping, learning rate scheduling, and dynamic batching when fine-tuning LLMs. \textbf{VII. Deployment:} We tested each model once, as running two fifteen-gigabyte models simultaneously was not possible on the inference system. Model fine-tuning was conducted on NVIDIA A100 graphics processing units with 80 GB of memory. \textbf{VIII. Evaluation Metrics:} Recognizing the limits of traditional natural language processing metrics, we employed fourteen different metrics to ensure proper benchmarking.

Among the fine-tuned models, most struggled to significantly improve performance; however, Qwen2.5-7B excelled in handling mixed content (code and text), and we recommend this model for bioinformatics tasks. Qwen2.5-7B achieved up to an 82\% improvement in BLEU-4 scores compared to the baseline for PRSGPT and a 6\% improvement for BioStarsGPT. Human evaluation further underscores the practical applicability of these domain-adapted models in real-world scenarios. PRSGPT attained 61.9\% accuracy in tool comparison tasks—comparable to Google Gemini's 61.4\%—while providing more detailed methodologies and more accurate citations, which are particularly valuable in scientific contexts. BioStarsGPT was evaluated on 142 questions and demonstrated a conceptual accuracy of 59\%. While its responses were precise, many required additional technical refinement.

During human testing, we faced multiple challenges, which would be useful for researchers developing similar LLMs. \textbf{I. Query Sensitivity (PRSGPT):} PRSGPT performed best with structured queries that included exact tool names and parameter references but struggled with comparative questions. This is obvious, as the dataset it was trained on did not contain examples of comparative questions comparing tools, and the fine-tuned model was unable to extract data from individual answers it was trained on to generate the comparative analysis. \textbf{II. Formatting Sensitivity (PRSGPT):} The model showed sensitivity to case variations and similar tool names (for example, LDPred versus LDPred2), occasionally leading to incorrect or mixed responses. \textbf{III. Citation Degradation (PRSGPT):} Citation and source identification accuracy declined under quantized versions of the model. \textbf{IV. Conceptual Breadth (BioStarsGPT):} BioStarsGPT performed well on conceptual and "how-to" questions related to workflows but lacked depth on technical queries involving specific parameters. \textbf{V. Inconsistent Detail (BioStarsGPT):} The model produced short and variable-length responses due to limitations in training data and struggled with multi-step procedural guidance. \textbf{VII. Common Failure Modes:} Both models struggled with out-of-domain queries, occasionally hallucinated when uncertain, and produced truncated responses for long inputs. \textbf{VIII. Feedback-Driven Enhancements:} An interesting observation was that when prompted for a detailed pipeline, BioStarsGPT primarily provided a directional framework. We then leveraged Google Gemini version 2.5 to augment this output, resulting in more comprehensive and enhanced responses, effectively expanding the knowledge base.

Despite promising outcomes, challenges remain. The use of single-turn question-answer pairs can limit the conversational depth required for complex problem-solving \cite{limitation1,limit2}. Additionally, large-scale training requires substantial computational resources; our seven-day window restricted full convergence for larger models.

This work provides empirical evidence that domain-specific fine-tuning enhances bioinformatics large language model performance. The open-source release of datasets and models enables community-driven improvements, serving as a blueprint for developing specialized scientific artificial intelligence assistants. Future directions include advancing domain-specific evaluation metrics, integrating retrieval-based literature access, and fostering collaborative model refinement. As bioinformatics workflows grow more complex, domain-trained large language models will become essential tools for accelerating discovery while maintaining scientific rigor \cite{Chen2025}.

\begin{mdframed}[linewidth=1pt,linecolor=black,
innerleftmargin=8pt,innerrightmargin=8pt,
innertopmargin=16pt-8.2pt,innerbottommargin=6pt]
{\fontsize{8.2pt}{10pt}\bfseries Key Points\par}
\begin{adjustwidth}{8pt}{0cm}
    \begin{itemize}
    \item We present a comprehensive 9-step pipeline for fine-tuning large language models on bioinformatics data, demonstrated through PRSGPT (PRS tools) and BioStarsGPT (forum discussions).
    \item The pipeline generated over 28,000 QA pairs for PRSGPT from diverse sources (PDFs, GitHub repositories, research articles) and 154,282 QA pairs for BioStarsGPT from community forum discussions.
    \item Qwen2.5-7B emerged as the best-performing model, achieving BLEU-4 improvements of 82\% for PRSGPT and 6\% for BioStarsGPT, with ROUGE-1 improvements of 70\% and 18\% respectively.
    \item The methodology integrates automated QA generation via Google Gemini, NLI for quality control, semantic deduplication, and parameter-efficient fine-tuning using LoRA, providing a scalable framework for domain-specific LLM adaptation.
    \end{itemize}
\end{adjustwidth}
\end{mdframed}

\clearpage

\section{Competing interests}
The authors declare that they have no competing interests

\section{Author contributions statement}
M.M. conceived the study, designed and implemented the pipeline, performed all analyses, generated datasets, and wrote the manuscript. D.B.A. supervised the project, contributed to experimental design, and reviewed the manuscript. Both authors approved the final version.

\section{Data availability}
The code for both \textbf{PRSGPT} and \textbf{BioStarsGPT} is publicly available on GitHub at \url{https://github.com/MuhammadMuneeb007/PolygenicRiskScoresGPT} and \url{https://github.com/MuhammadMuneeb007/BioStarsGPT}, respectively. The corresponding datasets can be accessed via Hugging Face at \url{https://huggingface.co/datasets/muhammadmuneeb007/PolygenicRiskScoresGPTDataset} and \url{https://huggingface.co/datasets/muhammadmuneeb007/BioStarsDataset}. The trained models are also hosted on Hugging Face at \url{https://huggingface.co/muhammadmuneeb007/PolygenicRiskScoresGPT} and \url{https://huggingface.co/muhammadmuneeb007/BioStarsGPT}, supporting full reproducibility and facilitating community-driven benchmarking.

\section{Acknowledgments}
Not applicable
\bibliographystyle{unsrt}
\bibliography{reference}

@article{chakraborty2023artificial,
  title={Artificial intelligence enabled ChatGPT and large language models in drug target discovery, drug discovery, and development},
  author={Chakraborty, Chiranjib and Bhattacharya, Manojit and Lee, Sang-Soo},
  journal={Molecular Therapy-Nucleic Acids},
  volume={33},
  pages={866--868},
  year={2023},
  publisher={Elsevier}
}

@article{Huang_2025, title={Methodologies and Advances in Fine-Tuning Techniques for Large Language Models}, url={http://dx.doi.org/10.36227/techrxiv.174682673.38816623/v1}, DOI={10.36227/techrxiv.174682673.38816623/v1}, publisher={Institute of Electrical and Electronics Engineers (IEEE)}, author={Huang, Zhu and Luo, Hongying and Wen, Yizhu and Wang, Tianyang and Song, Junhao and Bi, Ziqian and Liu, Ming and Song, Xinyuan and Hao, Junfeng and Liang, Chia Xin}, year={2025}, month=may }

@misc{Xu_2023,
  doi = {10.48550/ARXIV.2312.12148},
  url = {https://arxiv.org/abs/2312.12148},
  author = {Xu,  Lingling and Xie,  Haoran and Qin,  Si-Zhao Joe and Tao,  Xiaohui and Wang,  Fu Lee},
  title = {Parameter-Efficient Fine-Tuning Methods for Pretrained Language Models: A Critical Review and Assessment},
  publisher = {arXiv},
  year = {2023},
  copyright = {Creative Commons Attribution Non Commercial Share Alike 4.0 International}
}

@article{Zhang_2024, title={Unified Efficient Fine-Tuning Techniques for Open-Source Large Language Models}, url={http://dx.doi.org/10.21203/rs.3.rs-4660140/v1}, DOI={10.21203/rs.3.rs-4660140/v1}, publisher={Springer Science and Business Media LLC}, author={Zhang, Yulin and Li, Yanhua and Liu, Junhan}, year={2024}, month=jul }

@misc{Tian_2024,
  doi = {10.48550/ARXIV.2404.19245},
  url = {https://arxiv.org/abs/2404.19245},
  author = {Tian,  Chunlin and Shi,  Zhan and Guo,  Zhijiang and Li,  Li and Xu,  Chengzhong},
  title = {HydraLoRA: An Asymmetric LoRA Architecture for Efficient Fine-Tuning},
  publisher = {arXiv},
  year = {2024},
  copyright = {arXiv.org perpetual,  non-exclusive license}
}

@article{Shahid_2023, title={Leveraging Fine-Tuned Large Language Models in Bioinformatics: A Research Perspective}, ISSN={2632-3834}, url={http://dx.doi.org/10.32388/we7umn.2}, DOI={10.32388/we7umn.2}, journal={Qeios}, publisher={Qeios Ltd}, author={Shahid, Usama}, year={2023}, month=jul }

@article{Schmirler_2024, title={Fine-tuning protein language models boosts predictions across diverse tasks}, volume={15}, ISSN={2041-1723}, url={http://dx.doi.org/10.1038/s41467-024-51844-2}, DOI={10.1038/s41467-024-51844-2}, number={1}, journal={Nature Communications}, publisher={Springer Science and Business Media LLC}, author={Schmirler, Robert and Heinzinger, Michael and Rost, Burkhard}, year={2024}, month=aug }

@article{He2024,
  title = {Parameter-Efficient Fine-Tuning Enhances Adaptation of Single Cell Large Language Model for Cell Type Identification},
  url = {http://dx.doi.org/10.1101/2024.01.27.577455},
  DOI = {10.1101/2024.01.27.577455},
  publisher = {Cold Spring Harbor Laboratory},
  author = {He,  Fei and Fei,  Ruixin and Gao,  Mingyue and Su,  Li and Zhang,  Xinyu and Xu,  Dong},
  year = {2024},
  month = jan 
}

@misc{pile,
  doi = {10.48550/ARXIV.2101.00027},
  url = {https://arxiv.org/abs/2101.00027},
  author = {Gao,  Leo and Biderman,  Stella and Black,  Sid and Golding,  Laurence and Hoppe,  Travis and Foster,  Charles and Phang,  Jason and He,  Horace and Thite,  Anish and Nabeshima,  Noa and Presser,  Shawn and Leahy,  Connor},
  title = {The Pile: An 800GB Dataset of Diverse Text for Language Modeling},
  publisher = {arXiv},
  year = {2021},
  copyright = {Creative Commons Attribution 4.0 International}
}

@article{Parnell2011,
  title = {BioStar: An Online Question \&amp; Answer Resource for the Bioinformatics Community},
  volume = {7},
  ISSN = {1553-7358},
  url = {http://dx.doi.org/10.1371/journal.pcbi.1002216},
  DOI = {10.1371/journal.pcbi.1002216},
  number = {10},
  journal = {PLoS Computational Biology},
  publisher = {Public Library of Science (PLoS)},
  author = {Parnell,  Laurence D. and Lindenbaum,  Pierre and Shameer,  Khader and Dall’Olio,  Giovanni Marco and Swan,  Daniel C. and Jensen,  Lars Juhl and Cockell,  Simon J. and Pedersen,  Brent S. and Mangan,  Mary E. and Miller,  Christopher A. and Albert,  Istvan},
  editor = {Bourne,  Philip E.},
  year = {2011},
  month = oct,
  pages = {e1002216}
}

@article{Dorfner_2025, title={Evaluating the effectiveness of biomedical fine-tuning for large language models on clinical tasks}, volume={32}, ISSN={1527-974X}, url={http://dx.doi.org/10.1093/jamia/ocaf045}, DOI={10.1093/jamia/ocaf045}, number={6}, journal={Journal of the American Medical Informatics Association}, publisher={Oxford University Press (OUP)}, author={Dorfner, Felix J and Dada, Amin and Busch, Felix and Makowski, Marcus R and Han, Tianyu and Truhn, Daniel and Kleesiek, Jens and Sushil, Madhumita and Adams, Lisa C and Bressem, Keno K}, year={2025}, month=apr, pages={1015–1024} }

@inproceedings{_ajewska_2024, series={SIGIR-AP 2024}, title={Can Users Detect Biases or Factual Errors in Generated Responses in Conversational Information-Seeking?}, url={http://dx.doi.org/10.1145/3673791.3698409}, DOI={10.1145/3673791.3698409}, booktitle={Proceedings of the 2024 Annual International ACM SIGIR Conference on Research and Development in Information Retrieval in the Asia Pacific Region}, publisher={ACM}, author={Łajewska, Weronika and Balog, Krisztian and Spina, Damiano and Trippas, Johanne}, year={2024}, month=dec, pages={92–102}, collection={SIGIR-AP 2024} }

@article{Lamurias_2020, title={Generating Biomedical Question Answering Corpora From Q\&amp;A Forums}, volume={8}, ISSN={2169-3536}, url={http://dx.doi.org/10.1109/ACCESS.2020.3020868}, DOI={10.1109/access.2020.3020868}, journal={IEEE Access}, publisher={Institute of Electrical and Electronics Engineers (IEEE)}, author={Lamurias, Andre and Sousa, Diana and Couto, Francisco M.}, year={2020}, pages={161042–161051} }

@article{Ben_Abacha_2019, title={A question-entailment approach to question answering}, volume={20}, ISSN={1471-2105}, url={http://dx.doi.org/10.1186/s12859-019-3119-4}, DOI={10.1186/s12859-019-3119-4}, number={1}, journal={BMC Bioinformatics}, publisher={Springer Science and Business Media LLC}, author={Ben Abacha, Asma and Demner-Fushman, Dina}, year={2019}, month=oct }

@article{A_Moftah_2018, title={Performance Evaluation of Structured and Semi-Structured Bioinformatics Tools: A Comparative Study}, volume={9}, ISSN={0975-9018}, url={http://dx.doi.org/10.5121/ijsea.2018.9503}, DOI={10.5121/ijsea.2018.9503}, number={5}, journal={International Journal of Software Engineering \&amp; Applications}, publisher={Academy and Industry Research Collaboration Center (AIRCC)}, author={A. Moftah, Raja and M. Maatuk, Abdelsalam and White, Richard}, year={2018}, month=sep, pages={27–42} }

@inproceedings{Shen_2024, title={A Fine-tuning Dataset and Benchmark for Large Language Models for Protein Understanding}, url={http://dx.doi.org/10.1109/bibm62325.2024.10821894}, DOI={10.1109/bibm62325.2024.10821894}, booktitle={2024 IEEE International Conference on Bioinformatics and Biomedicine (BIBM)}, publisher={IEEE}, author={Shen, Yiqing and Chen, Zan and Mamalakis, Michail and He, Luhan and Xia, Haiyang and Li, Tianbin and Su, Yanzhou and He, Junjun and Wang, Yu Guang}, year={2024}, month=dec, pages={2390–2395} }

@article{Sledzieski_2024, title={Democratizing protein language models with parameter-efficient fine-tuning}, volume={121}, ISSN={1091-6490}, url={http://dx.doi.org/10.1073/pnas.2405840121}, DOI={10.1073/pnas.2405840121}, number={26}, journal={Proceedings of the National Academy of Sciences}, publisher={Proceedings of the National Academy of Sciences}, author={Sledzieski, Samuel and Kshirsagar, Meghana and Baek, Minkyung and Dodhia, Rahul and Lavista Ferres, Juan and Berger, Bonnie}, year={2024}, month=jun }

@inproceedings{Wu_2024, title={Mixture-of-Skills: Learning to Optimize Data Usage for Fine-Tuning Large Language Models}, url={http://dx.doi.org/10.18653/v1/2024.emnlp-main.787}, DOI={10.18653/v1/2024.emnlp-main.787}, booktitle={Proceedings of the 2024 Conference on Empirical Methods in Natural Language Processing}, publisher={Association for Computational Linguistics}, author={Wu, Minghao and Vu, Thuy-Trang and Qu, Lizhen and Haf, Reza}, year={2024}, pages={14226–14240} }

@article{Tinn_2023, title={Fine-tuning large neural language models for biomedical natural language processing}, volume={4}, ISSN={2666-3899}, url={http://dx.doi.org/10.1016/j.patter.2023.100729}, DOI={10.1016/j.patter.2023.100729}, number={4}, journal={Patterns}, publisher={Elsevier BV}, author={Tinn, Robert and Cheng, Hao and Gu, Yu and Usuyama, Naoto and Liu, Xiaodong and Naumann, Tristan and Gao, Jianfeng and Poon, Hoifung}, year={2023}, month=apr, pages={100729} }

@inproceedings{lin-2004-rouge,
    title = "{ROUGE}: A Package for Automatic Evaluation of Summaries",
    author = "Lin, Chin-Yew",
    booktitle = "Text Summarization Branches Out",
    month = jul,
    year = "2004",
    address = "Barcelona, Spain",
    publisher = "Association for Computational Linguistics",
    url = "https://aclanthology.org/W04-1013/",
    pages = "74--81"
}

@inproceedings{papineni-etal-2002-bleu,
    title = "{B}leu: a Method for Automatic Evaluation of Machine Translation",
    author = "Papineni, Kishore and
              Roukos, Salim and
              Ward, Todd and
              Zhu, Wei-Jing",
    editor = "Isabelle, Pierre and
              Charniak, Eugene and
              Lin, Dekang",
    booktitle = "Proceedings of the 40th Annual Meeting of the Association for Computational Linguistics",
    month = jul,
    year = "2002",
    address = "Philadelphia, Pennsylvania, USA",
    publisher = "Association for Computational Linguistics",
    url = "https://aclanthology.org/P02-1040/",
    doi = "10.3115/1073083.1073135",
    pages = "311--318"
}

@inproceedings{banerjee-lavie-2005-meteor,
    title = "{METEOR}: An Automatic Metric for {MT} Evaluation with Improved Correlation with Human Judgments",
    author = "Banerjee, Satanjeev and
              Lavie, Alon",
    editor = "Goldstein, Jade and
              Lavie, Alon and
              Lin, Chin-Yew and
              Voss, Clare",
    booktitle = "Proceedings of the {ACL} Workshop on Intrinsic and Extrinsic Evaluation Measures for Machine Translation and/or Summarization",
    month = jun,
    year = "2005",
    address = "Ann Arbor, Michigan",
    publisher = "Association for Computational Linguistics",
    url = "https://aclanthology.org/W05-0909/",
    pages = "65--72"
}

@article{levenshtein1966binary,
  title={Binary codes capable of correcting deletions, insertions, and reversals},
  author={Levenshtein, Vladimir Iosifovich},
  journal={Soviet Physics Doklady},
  volume={10},
  number={8},
  pages={707--710},
  year={1966},
  note={English translation of the original 1965 Russian paper}
}

@article{jaccard1901etude,
  title={Étude comparative de la distribution florale dans une portion des Alpes et des Jura},
  author={Jaccard, Paul},
  journal={Bulletin de la Société Vaudoise des Sciences Naturelles},
  volume={37},
  number={142},
  pages={547--579},
  year={1901},
  doi={10.5169/seals-266450},
  note={Original paper introducing the Jaccard coefficient (coefficient de communauté)}
}

@article{sparckjones1972statistical,
  title={A statistical interpretation of term specificity and its application in retrieval},
  author={Spärck Jones, Karen},
  journal={Journal of Documentation},
  volume={28},
  number={1},
  pages={11--21},
  year={1972},
  doi={10.1108/eb026526},
  note={Seminal paper introducing inverse document frequency (IDF)}
}

@article{salton1988term,
  title={Term-weighting approaches in automatic text retrieval},
  author={Salton, Gerard and Buckley, Christopher},
  journal={Information Processing \& Management},
  volume={24},
  number={5},
  pages={513--523},
  year={1988},
  doi={10.1016/0306-4573(88)90021-0},
  note={Comprehensive treatment of TF-IDF and related term weighting schemes}
}

@inproceedings{reimers-gurevych-2019-sentence,
    title = "Sentence-{BERT}: Sentence Embeddings using {S}iamese {BERT}-Networks",
    author = "Reimers, Nils and Gurevych, Iryna",
    editor = "Inui, Kentaro and Jiang, Jing and Ng, Vincent and Wan, Xiaojun",
    booktitle = "Proceedings of the 2019 Conference on Empirical Methods in Natural Language Processing and the 9th International Joint Conference on Natural Language Processing (EMNLP-IJCNLP)",
    month = nov,
    year = "2019",
    address = "Hong Kong, China",
    publisher = "Association for Computational Linguistics",
    url = "https://aclanthology.org/D19-1410/",
    doi = "10.18653/v1/D19-1410",
    pages = "3982--3992"
}

@misc{honnibal2020spacy,
    title = {{spaCy: Industrial-strength Natural Language Processing in Python}},
    author = {Honnibal, Matthew and 
              Montani, Ines and 
              Van Landeghem, Sofie and 
              Boyd, Adriane},
    year = {2020},
    doi = {10.5281/zenodo.1212303},
    url = {https://spacy.io}
}

@inproceedings{kusner2015word,
  title={From Word Embeddings To Document Distances},
  author={Kusner, Matt J. and Sun, Yu and Kolkin, Nicholas I. and Weinberger, Kilian Q.},
  booktitle={Proceedings of the 32nd International Conference on Machine Learning},
  pages={957--966},
  year={2015},
  volume={37},
  series={Proceedings of Machine Learning Research},
  publisher={PMLR},
  address={Lille, France},
  month={07--09 Jul},
  url={http://proceedings.mlr.press/v37/kusnerb15.pdf},
  note={Original paper introducing Word Mover's Distance}
}

@inproceedings{zhang-etal-2020-bertscore,
    title = "{BERT}Score: Evaluating Text Generation with {BERT}",
    author = "Zhang, Tianyi and Kishore, Varsha and Wu, Felix and Weinberger, Kilian Q. and Artzi, Yoav",
    booktitle = "International Conference on Learning Representations",
    year = "2020",
    url = "https://openreview.net/forum?id=SkeHuCVFDr"
}

@inproceedings{feng-etal-2020-codebert,
    title = "{C}ode{BERT}: A Pre-Trained Model for Programming and Natural Languages",
    author = "Feng, Zhangyin and Guo, Daya and Tang, Duyu and Duan, Nan and Feng, Xiaocheng and Gong, Ming and Shou, Linjun and Qin, Bing and Liu, Ting and Jiang, Daxin and Zhou, Ming",
    editor = "Cohn, Trevor and He, Yulan and Liu, Yang",
    booktitle = "Findings of the Association for Computational Linguistics: EMNLP 2020",
    month = nov,
    year = "2020",
    address = "Online",
    publisher = "Association for Computational Linguistics",
    url = "https://aclanthology.org/2020.findings-emnlp.139/",
    doi = "10.18653/v1/2020.findings-emnlp.139",
    pages = "1536--1547"
}

@inproceedings{bowman-etal-2015-large,
  title={A large annotated corpus for learning natural language inference},
  author={Bowman, Samuel R. and Angeli, Gabor and Potts, Christopher and Manning, Christopher D.},
  editor={Màrquez, Lluís and Callison-Burch, Chris and Su, Jian},
  booktitle={Proceedings of the 2015 Conference on Empirical Methods in Natural Language Processing},
  month={sep},
  year={2015},
  address={Lisbon, Portugal},
  publisher={Association for Computational Linguistics},
  url={https://aclanthology.org/D15-1075/},
  doi={10.18653/v1/D15-1075},
  pages={632--642}
}

@inproceedings{devlin-etal-2019-bert,
  title={{BERT}: Pre-training of Deep Bidirectional Transformers for Language Understanding},
  author={Devlin, Jacob and Chang, Ming-Wei and Lee, Kenton and Toutanova, Kristina},
  booktitle={Proceedings of the 2019 Conference of the North {A}merican Chapter of the Association for Computational Linguistics: Human Language Technologies, Volume 1 (Long and Short Papers)},
  month={jun},
  year={2019},
  address={Minneapolis, Minnesota},
  publisher={Association for Computational Linguistics},
  url={https://aclanthology.org/N19-1423/},
  doi={10.18653/v1/N19-1423},
  pages={4171--4186}
}

@article{neyman1934two,
  title={On the Two Different Aspects of the Representative Method: The Method of Stratified Sampling and the Method of Purposive Selection},
  author={Neyman, Jerzy},
  journal={Journal of the Royal Statistical Society},
  volume={97},
  number={4},
  pages={558--606},
  year={1934},
  publisher={Royal Statistical Society},
  doi={10.1111/j.2397-2335.1934.tb04184.x}
}

@article{touvron2023llama,
  title={LLaMA: Open and Efficient Foundation Language Models},
  author={Touvron, Hugo and Lavril, Thibaut and Izacard, Gautier and Martinet, Xavier and Lachaux, Marie-Anne and Lacroix, Timothée and Rozière, Baptiste and Goyal, Naman and Hambro, Eric and Azhar, Faisal and Rodriguez, Aurelien and Joulin, Armand and Grave, Edouard and Lample, Guillaume},
  journal={arXiv preprint arXiv:2302.13971},
  year={2023},
  doi={10.48550/arXiv.2302.13971}
}

@article{bai2023qwen,
  title={Qwen Technical Report},
  author={Bai, Jinze and Bai, Shuai and Chu, Yunfei and Cui, Zeyu and Dang, Kai and Deng, Xiaodong and Fan, Yang and Ge, Wenbin and Han, Yu and Huang, Fei and Hui, Binyuan and Ji, Luo and Li, Mei and Lin, Junyang and Lin, Runji and Liu, Dayiheng and Liu, Gao and Lu, Chengqiang and Lu, Keming and Ma, Jianxin and Men, Rui and Ren, Xingzhang and Ren, Xuancheng and Tan, Chuanqi and Tan, Sinan and Tu, Jianhong and Wang, Peng and Wang, Shijie and Wang, Wei and Wu, Shengguang and Xu, Benfeng and Xu, Jin and Yang, An and Yang, Hao and Yang, Jian and Yang, Shusheng and Yao, Yang and Yu, Bowen and Yuan, Hongyi and Yuan, Zheng and Zhang, Jianwei and Zhang, Xingxuan and Zhang, Yichang and Zhang, Zhenru and Zhou, Chang and Zhou, Jingren and Zhou, Xiaohuan and Zhu, Tianhang},
  journal={arXiv preprint arXiv:2309.16609},
  year={2023},
  doi={10.48550/arXiv.2309.16609}
}

@article{mesnard2024gemma,
  title={Gemma: Open Models Based on Gemini Research and Technology},
  author={{Gemma Team}},
  journal={arXiv preprint arXiv:2403.08295},
  year={2024},
  doi={10.48550/arXiv.2403.08295}
}

@article{gunasekar2023textbooks,
  title={Textbooks Are All You Need},
  author={Gunasekar, Suriya and Zhang, Yi and Aneja, Jyoti and Mendes, Caio C. T. and Del Giorno, Allie and Gopi, Sivakanth and Javaheripi, Mojan and Kauffmann, Piero and de Rosa, Gustavo and Saarikivi, Olli and Salim, Adil and Shah, Shital and Behl, Harkirat Singh and Wang, Xin and Bubeck, Sébastien and Eldan, Ronen and Kalai, Adam Tauman and Lee, Yin Tat and Li, Yuanzhi},
  journal={arXiv preprint arXiv:2306.11644},
  year={2023},
  doi={10.48550/arXiv.2306.11644}
}

@article{jiang2023mistral,
  title={Mistral 7B},
  author={Jiang, Albert Q. and Sablayrolles, Alexandre and Mensch, Arthur and Bamford, Chris and Chaplot, Devendra Singh and de las Casas, Diego and Bressand, Florian and Lengyel, Gianna and Lample, Guillaume and Saulnier, Lucile and Lavaud, Lélio Renard and Lachaux, Marie-Anne and Stock, Pierre and Le Scao, Teven and Lavril, Thibaut and Wang, Thomas and Lacroix, Timothée and El Sayed, William},
  journal={arXiv preprint arXiv:2310.06825},
  year={2023},
  doi={10.48550/arXiv.2310.06825}
}

@inproceedings{hu2022lora,
  title={{LoRA}: Low-Rank Adaptation of Large Language Models},
  author={Edward J. Hu and Yelong Shen and Phillip Wallis and Zeyuan Allen-Zhu and Yuanzhi Li and Shean Wang and Lu Wang and Weizhu Chen},
  booktitle={International Conference on Learning Representations},
  year={2022},
  url={https://openreview.net/forum?id=nZeVKeeFYf9},
  note={ICLR 2022}
}

@article{Chen2025,
  title = {Benchmarking large language models for biomedical natural language processing applications and recommendations},
  volume = {16},
  ISSN = {2041-1723},
  url = {http://dx.doi.org/10.1038/s41467-025-56989-2},
  DOI = {10.1038/s41467-025-56989-2},
  number = {1},
  journal = {Nature Communications},
  publisher = {Springer Science and Business Media LLC},
  author = {Chen,  Qingyu and Hu,  Yan and Peng,  Xueqing and Xie,  Qianqian and Jin,  Qiao and Gilson,  Aidan and Singer,  Maxwell B. and Ai,  Xuguang and Lai,  Po-Ting and Wang,  Zhizheng and Keloth,  Vipina K. and Raja,  Kalpana and Huang,  Jimin and He,  Huan and Lin,  Fongci and Du,  Jingcheng and Zhang,  Rui and Zheng,  W. Jim and Adelman,  Ron A. and Lu,  Zhiyong and Xu,  Hua},
  year = {2025},
  month = apr 
}

@article{Simon2024,
  title = {Language models for biological research: a primer},
  volume = {21},
  ISSN = {1548-7105},
  url = {http://dx.doi.org/10.1038/s41592-024-02354-y},
  DOI = {10.1038/s41592-024-02354-y},
  number = {8},
  journal = {Nature Methods},
  publisher = {Springer Science and Business Media LLC},
  author = {Simon,  Elana and Swanson,  Kyle and Zou,  James},
  year = {2024},
  month = aug,
  pages = {1422–1429}
}

@article{dettmers2023qlora,
  title={{QLoRA}: Efficient Finetuning of Quantized {LLM}s},
  author={Tim Dettmers and Artidoro Pagnoni and Ari Holtzman and Luke Zettlemoyer},
  journal={arXiv preprint arXiv:2305.14314},
  year={2023},
  url={https://arxiv.org/abs/2305.14314}
}

@article{Li2024,
  title = {Progress and opportunities of foundation models in bioinformatics},
  volume = {25},
  ISSN = {1477-4054},
  url = {http://dx.doi.org/10.1093/bib/bbae548},
  DOI = {10.1093/bib/bbae548},
  number = {6},
  journal = {Briefings in Bioinformatics},
  publisher = {Oxford University Press (OUP)},
  author = {Li,  Qing and Hu,  Zhihang and Wang,  Yixuan and Li,  Lei and Fan,  Yimin and King,  Irwin and Jia,  Gengjie and Wang,  Sheng and Song,  Le and Li,  Yu},
  year = {2024},
  month = sep 
}

@software{unsloth,
  author = {Daniel Han and Michael Han and Unsloth team},
  title = {Unsloth},
  url = {https://github.com/unslothai/unsloth},
  year = {2023}
}

@inproceedings{wolf-etal-2020-transformers,
  title = "Transformers: State-of-the-Art Natural Language Processing",
  author = "Wolf, Thomas and
            Debut, Lysandre and
            Sanh, Victor and
            Chaumond, Julien and
            Delangue, Clement and
            Moi, Anthony and
            Cistac, Pierric and
            Rault, Tim and
            Louf, Remi and
            Funtowicz, Morgan and
            Davison, Joe and
            Shleifer, Sam and
            von Platen, Patrick and
            Ma, Clara and
            Jernite, Yacine and
            Plu, Julien and
            Xu, Canwen and
            Le Scao, Teven and
            Gugger, Sylvain and
            Drame, Mariama and
            Lhoest, Quentin and
            Rush, Alexander",
  editor = "Liu, Qun and
            Schlangen, David",
  booktitle = "Proceedings of the 2020 Conference on Empirical Methods in Natural Language Processing: System Demonstrations",
  month = oct,
  year = "2020",
  address = "Online",
  publisher = "Association for Computational Linguistics",
  url = "https://aclanthology.org/2020.emnlp-demos.6/",
  doi = "10.18653/v1/2020.emnlp-demos.6",
  pages = "38--45"
}

@inproceedings{schleimer2003winnowing,
    title = "Winnowing: local algorithms for document fingerprinting",
    author = "Schleimer, Saul and 
              Wilkerson, Daniel S. and 
              Aiken, Alex",
    booktitle = "Proceedings of the 2003 ACM SIGMOD International Conference on Management of Data",
    year = "2003",
    address = "San Diego, California, USA",
    publisher = "Association for Computing Machinery",
    pages = "76--85",
    doi = "10.1145/872757.872770"
}

@article{salton1975vector,
  title   = {A Vector Space Model for Automatic Indexing},
  author  = {Salton, Gerard and Wong, A. and Yang, C. S.},
  journal = {Communications of the ACM},
  volume  = {18},
  number  = {11},
  pages   = {613--620},
  year    = {1975},
  doi     = {10.1145/361219.361220}
}

@article{Piantadosi2014,
  title = {Zipf’s word frequency law in natural language: A critical review and future directions},
  volume = {21},
  ISSN = {1531-5320},
  url = {http://dx.doi.org/10.3758/s13423-014-0585-6},
  DOI = {10.3758/s13423-014-0585-6},
  number = {5},
  journal = {Psychonomic Bulletin \&amp; Review},
  publisher = {Springer Science and Business Media LLC},
  author = {Piantadosi,  Steven T.},
  year = {2014},
  month = mar,
  pages = {1112–1130}
}

@article{pearson1896mathematical,
  title={Mathematical Contributions to the Theory of Evolution. {III}. {Regression}, Heredity and Panmixia},
  author={Pearson, Karl},
  journal={Philosophical Transactions of the Royal Society of London},
  volume={187},
  pages={253--318},
  year={1896},
  publisher={Royal Society},
  doi={10.1098/rsta.1896.0007},
  note={First rigorous mathematical treatment of correlation coefficient}
}

@article{vandermaaten08a,
  author  = {Laurens van der Maaten and Geoffrey Hinton},
  title   = {Visualizing Data using t-SNE},
  journal = {Journal of Machine Learning Research},
  year    = {2008},
  volume  = {9},
  pages   = {2579-2605},
  url     = {http://www.jmlr.org/papers/v9/vandermaaten08a.html}
}

@article{Wang_2024, title={Fine-tuning large language models for rare disease concept normalization}, volume={31}, ISSN={1527-974X}, url={http://dx.doi.org/10.1093/jamia/ocae133}, DOI={10.1093/jamia/ocae133}, number={9}, journal={Journal of the American Medical Informatics Association}, publisher={Oxford University Press (OUP)}, author={Wang, Andy and Liu, Cong and Yang, Jingye and Weng, Chunhua}, year={2024}, month=jun, pages={2076–2083} }

@article{wei2022chain,
  title={Chain-of-Thought Prompting Elicits Reasoning in Large Language Models},
  author={Wei, Jason and Wang, Xuezhi and Schuurmans, Dale and Bosma, Maarten and Ichter, Brian and Xia, Fei and Chi, Ed and Le, Quoc and Zhou, Denny},
  journal={arXiv preprint arXiv:2201.11903},
  year={2022},
  url={https://arxiv.org/abs/2201.11903}
}

@book{tukey1977exploratory,
  title     = {Exploratory Data Analysis},
  author    = {Tukey, John W.},
  year      = {1977},
  publisher = {Addison-Wesley},
  isbn      = {9780201076165}
}

@article{dhillon2001concept,
  title={Concept decompositions for large sparse text data using clustering},
  author={Dhillon, Inderjit S. and Modha, Dharmendra S.},
  journal={Machine Learning},
  volume={42},
  number={1-2},
  pages={143--175},
  year={2001},
  publisher={Springer},
  doi={10.1023/A:1007612920971}
}

@article{Madani_2023, title={Large language models generate functional protein sequences across diverse families}, volume={41}, ISSN={1546-1696}, url={http://dx.doi.org/10.1038/s41587-022-01618-2}, DOI={10.1038/s41587-022-01618-2}, number={8}, journal={Nature Biotechnology}, publisher={Springer Science and Business Media LLC}, author={Madani, Ali and Krause, Ben and Greene, Eric R. and Subramanian, Subu and Mohr, Benjamin P. and Holton, James M. and Olmos, Jose Luis and Xiong, Caiming and Sun, Zachary Z. and Socher, Richard and Fraser, James S. and Naik, Nikhil}, year={2023}, month=jan, pages={1099–1106} }

@article{Fang_2024, title={Automated Federated Pipeline for Parameter-Efficient Fine-Tuning of Large Language Models}, url={http://dx.doi.org/10.36227/techrxiv.173272996.63900291/v1}, DOI={10.36227/techrxiv.173272996.63900291/v1}, publisher={Institute of Electrical and Electronics Engineers (IEEE)}, author={Fang, Zihan and Lin, Zheng and Chen, Zhe and Chen, Xianhao and Gao, Yue and Fang, Yuguang}, year={2024}, month=nov }

@phdthesis{Chavan, title={Large Language Models for Bacterial Genomic Analysis}, url={http://dx.doi.org/10.31979/etd.573m-zvbd}, DOI={10.31979/etd.573m-zvbd}, school={San Jose State University Library}, author={Chavan, Manvendra} }

@article{Keloth2024,
  title = {Advancing entity recognition in biomedicine via instruction tuning of large language models},
  volume = {40},
  ISSN = {1367-4811},
  url = {http://dx.doi.org/10.1093/bioinformatics/btae163},
  DOI = {10.1093/bioinformatics/btae163},
  number = {4},
  journal = {Bioinformatics},
  publisher = {Oxford University Press (OUP)},
  author = {Keloth,  Vipina K and Hu,  Yan and Xie,  Qianqian and Peng,  Xueqing and Wang,  Yan and Zheng,  Andrew and Selek,  Melih and Raja,  Kalpana and Wei,  Chih Hsuan and Jin,  Qiao and Lu,  Zhiyong and Chen,  Qingyu and Xu,  Hua},
  editor = {Wren,  Jonathan},
  year = {2024},
  month = mar 
}

@article{Yang2020,
  title = {Measurement of Semantic Textual Similarity in Clinical Texts: Comparison of Transformer-Based Models},
  volume = {8},
  ISSN = {2291-9694},
  url = {http://dx.doi.org/10.2196/19735},
  DOI = {10.2196/19735},
  number = {11},
  journal = {JMIR Medical Informatics},
  publisher = {JMIR Publications Inc.},
  author = {Yang,  Xi and He,  Xing and Zhang,  Hansi and Ma,  Yinghan and Bian,  Jiang and Wu,  Yonghui},
  year = {2020},
  month = nov,
  pages = {e19735}
}

@misc{limitation1,
  doi = {10.48550/ARXIV.2505.06120},
  url = {https://arxiv.org/abs/2505.06120},
  author = {Laban,  Philippe and Hayashi,  Hiroaki and Zhou,  Yingbo and Neville,  Jennifer},
  title = {LLMs Get Lost In Multi-Turn Conversation},
  publisher = {arXiv},
  year = {2025},
  copyright = {Creative Commons Attribution 4.0 International}
}

@article{Anisuzzaman2025,
  title = {Fine-Tuning Large Language Models for Specialized Use Cases},
  volume = {3},
  ISSN = {2949-7612},
  url = {http://dx.doi.org/10.1016/j.mcpdig.2024.11.005},
  DOI = {10.1016/j.mcpdig.2024.11.005},
  number = {1},
  journal = {Mayo Clinic Proceedings: Digital Health},
  publisher = {Elsevier BV},
  author = {Anisuzzaman,  D.M. and Malins,  Jeffrey G. and Friedman,  Paul A. and Attia,  Zachi I.},
  year = {2025},
  month = mar,
  pages = {100184}
}

@article{Dessimoz2024,
  title = {AI and the democratization of knowledge},
  volume = {11},
  ISSN = {2052-4463},
  url = {http://dx.doi.org/10.1038/s41597-024-03099-1},
  DOI = {10.1038/s41597-024-03099-1},
  number = {1},
  journal = {Scientific Data},
  publisher = {Springer Science and Business Media LLC},
  author = {Dessimoz,  Christophe and Thomas,  Paul D.},
  year = {2024},
  month = mar 
}

@misc{limit2,
  doi = {10.48550/ARXIV.2308.08747},
  url = {https://arxiv.org/abs/2308.08747},
  author = {Luo,  Yun and Yang,  Zhen and Meng,  Fandong and Li,  Yafu and Zhou,  Jie and Zhang,  Yue},
  title = {An Empirical Study of Catastrophic Forgetting in Large Language Models During Continual Fine-tuning},
  publisher = {arXiv},
  year = {2023},
  copyright = {Creative Commons Attribution 4.0 International}
}

@misc{finetuning,
  doi = {10.48550/ARXIV.2408.13296},
  url = {https://arxiv.org/abs/2408.13296},
  author = {Parthasarathy,  Venkatesh Balavadhani and Zafar,  Ahtsham and Khan,  Aafaq and Shahid,  Arsalan},
  title = {The Ultimate Guide to Fine-Tuning LLMs from Basics to Breakthroughs: An Exhaustive Review of Technologies,  Research,  Best Practices,  Applied Research Challenges and Opportunities},
  publisher = {arXiv},
  year = {2024},
  copyright = {Creative Commons Attribution Non Commercial No Derivatives 4.0 International}
}

@article{Sarumi2024,
  title = {Large language models and their applications in bioinformatics},
  volume = {23},
  ISSN = {2001-0370},
  url = {http://dx.doi.org/10.1016/j.csbj.2024.09.031},
  DOI = {10.1016/j.csbj.2024.09.031},
  journal = {Computational and Structural Biotechnology Journal},
  publisher = {Elsevier BV},
  author = {Sarumi,  Oluwafemi A. and Heider,  Dominik},
  year = {2024},
  month = dec,
  pages = {3498–3505}
}

@article{liu2024large,
  title={Large language models in bioinformatics: applications and perspectives},
  author={Liu, Jiajia and Yang, Mengyuan and Yu, Yankai and Xu, Haixia and Li, Kang and Zhou, Xiaobo},
  journal={arXiv preprint arXiv:2401.04155},
  year={2024}
}

@inproceedings{nazi2024large,
  title={Large language models in healthcare and medical domain: A review},
  author={Nazi, Zabir Al and Peng, Wei},
  booktitle={Informatics},
  volume={11},
  number={3},
  pages={57},
  year={2024},
  organization={MDPI}
}

@article{green2025litsumm,
  title={LitSumm: large language models for literature summarization of noncoding RNAs},
  author={Green, Andrew and Ribas, Carlos Eduardo and Ontiveros-Palacios, Nancy and Griffiths-Jones, Sam and Petrov, Anton I and Bateman, Alex and Sweeney, Blake},
  journal={Database},
  volume={2025},
  pages={baaf006},
  year={2025},
  publisher={Oxford University Press UK}
}

@article{bednarczyk2025scientific,
  title={Scientific Evidence for Clinical Text Summarization Using Large Language Models: Scoping Review},
  author={Bednarczyk, Lydie and Reichenpfader, Daniel and Gaudet-Blavignac, Christophe and Ette, Amon Kenna and Zaghir, Jamil and Zheng, Yuanyuan and Bensahla, Adel and Bjelogrlic, Mina and Lovis, Christian},
  journal={Journal of Medical Internet Research},
  volume={27},
  pages={e68998},
  year={2025},
  publisher={JMIR Publications Toronto, Canada}
}

@incollection{sreenivasanlarge,
  title={Large Language Models in Healthcare Information Systems: Overcoming Challenges to Achieve Personalized Care},
  author={Sreenivasan, M and Narayan, Abhay},
  booktitle={Generative Intelligence in Healthcare},
  pages={125--149},
  publisher={CRC Press}
}

@article{lee2020biobert,
  title={BioBERT: a pre-trained biomedical language representation model for biomedical text mining},
  author={Lee, Jinhyuk and Yoon, Wonjin and Kim, Sungdong and Kim, Donghyeon and Kim, Sunkyu and So, Chan Ho and Kang, Jaewoo},
  journal={Bioinformatics},
  volume={36},
  number={4},
  pages={1234--1240},
  year={2020},
  publisher={Oxford University Press},
  doi={10.1093/bioinformatics/btz682}
}

@article{gu2021domain,
  title={Domain-specific language model pretraining for biomedical NLP},
  author={Gu, Yu and Tinn, Robert and Cheng, Hao and Lucas, Michael and Usuyama, Naoto and Liu, Xiaodong and Naumann, Tristan and Gao, Jianfeng and Poon, Hoifung},
  journal={BMC Bioinformatics},
  volume={22},
  number={1},
  pages={1--14},
  year={2021},
  publisher={BioMed Central},
  doi={10.1186/s12859-021-04211-4}
}

@article{Greene2014,
  title = {Big Data Bioinformatics},
  volume = {229},
  ISSN = {1097-4652},
  url = {http://dx.doi.org/10.1002/jcp.24662},
  DOI = {10.1002/jcp.24662},
  number = {12},
  journal = {Journal of Cellular Physiology},
  publisher = {Wiley},
  author = {Greene,  Casey S. and Tan,  Jie and Ung,  Matthew and Moore,  Jason H. and Cheng,  Chao},
  year = {2014},
  month = aug,
  pages = {1896–1900}
}

@article{Saparov2025,
  title = {Big data and transformative bioinformatics in genomic diagnostics and beyond},
  volume = {134},
  ISSN = {1353-8020},
  url = {http://dx.doi.org/10.1016/j.parkreldis.2025.107311},
  DOI = {10.1016/j.parkreldis.2025.107311},
  journal = {Parkinsonism \&amp; Related Disorders},
  publisher = {Elsevier BV},
  author = {Saparov,  Alice and Zech,  Michael},
  year = {2025},
  month = may,
  pages = {107311}
}

@ARTICLE{10433480,
  author={Raiaan, Mohaimenul Azam Khan and Mukta, Md. Saddam Hossain and Fatema, Kaniz and Fahad, Nur Mohammad and Sakib, Sadman and Mim, Most Marufatul Jannat and Ahmad, Jubaer and Ali, Mohammed Eunus and Azam, Sami},
  journal={IEEE Access}, 
  title={A Review on Large Language Models: Architectures, Applications, Taxonomies, Open Issues and Challenges}, 
  year={2024},
  volume={12},
  number={},
  pages={26839-26874},
  doi={10.1109/ACCESS.2024.3365742}}

@misc{llms,
  doi = {10.48550/ARXIV.2307.06435},
  url = {https://arxiv.org/abs/2307.06435},
  author = {Naveed,  Humza and Khan,  Asad Ullah and Qiu,  Shi and Saqib,  Muhammad and Anwar,  Saeed and Usman,  Muhammad and Akhtar,  Naveed and Barnes,  Nick and Mian,  Ajmal},
  title = {A Comprehensive Overview of Large Language Models},
  publisher = {arXiv},
  year = {2023},
  copyright = {Creative Commons Attribution 4.0 International}
}

@article{Hu2022,
  title = {Challenges in Bioinformatics Workflows for Processing Microbiome Omics Data at Scale},
  volume = {1},
  ISSN = {2673-7647},
  url = {http://dx.doi.org/10.3389/fbinf.2021.826370},
  DOI = {10.3389/fbinf.2021.826370},
  journal = {Frontiers in Bioinformatics},
  publisher = {Frontiers Media SA},
  author = {Hu,  Bin and Canon,  Shane and Eloe-Fadrosh,  Emiley A. and Anubhav and Babinski,  Michal and Corilo,  Yuri and Davenport,  Karen and Duncan,  William D. and Fagnan,  Kjiersten and Flynn,  Mark and Foster,  Brian and Hays,  David and Huntemann,  Marcel and Jackson,  Elais K. Player and Kelliher,  Julia and Li,  Po-E. and Lo,  Chien-Chi and Mans,  Douglas and McCue,  Lee Ann and Mouncey,  Nigel and Mungall,  Christopher J. and Piehowski,  Paul D. and Purvine,  Samuel O. and Smith,  Montana and Varghese,  Neha Jacob and Winston,  Donald and Xu,  Yan and Chain,  Patrick S. G.},
  year = {2022},
  month = jan 
}

@article{MorrisonSmith2022,
  title = {Challenges in large-scale bioinformatics projects},
  volume = {9},
  ISSN = {2662-9992},
  url = {http://dx.doi.org/10.1057/s41599-022-01141-4},
  DOI = {10.1057/s41599-022-01141-4},
  number = {1},
  journal = {Humanities and Social Sciences Communications},
  publisher = {Springer Science and Business Media LLC},
  author = {Morrison-Smith,  Sarah and Boucher,  Christina and Sarcevic,  Aleksandra and Noyes,  Noelle and O’Brien,  Catherine and Cuadros,  Nazaret and Ruiz,  Jaime},
  year = {2022},
  month = apr 
}

@article{Ma2021,
  title = {Genetic prediction of complex traits with polygenic scores: a statistical review},
  volume = {37},
  ISSN = {0168-9525},
  url = {http://dx.doi.org/10.1016/j.tig.2021.06.004},
  DOI = {10.1016/j.tig.2021.06.004},
  number = {11},
  journal = {Trends in Genetics},
  publisher = {Elsevier BV},
  author = {Ma,  Ying and Zhou,  Xiang},
  year = {2021},
  month = nov,
  pages = {995–1011}
}

\end{document}